\documentclass[10pt,twocolumn,letterpaper]{article}

\usepackage[pagenumbers]{iccv} % To force page numbers, e.g. for an arXiv version
\usepackage{multirow}

\usepackage{pifont}
\usepackage{indentfirst}
\definecolor{iccvblue}{rgb}{0.21,0.49,0.74}
\definecolor{darkgreen}{rgb}{0.0, 0.7, 0.0}
\definecolor{darkred}{rgb}{0.7, 0, 0}  % 
\usepackage[pagebackref,breaklinks,colorlinks,allcolors=iccvblue]{hyperref}

\def\paperID{7067} % *** Enter the Paper ID here
\def\confName{ICCV}
\def\confYear{2025}

\title{You Are Your Own Best Teacher: Achieving Centralized-level Performance in Federated Learning under Heterogeneous and Long-tailed Data}

\author{
    Shanshan Yan$^{1}$ \quad
    Zexi Li$^{2}$ \quad
    Chao Wu$^{3}$ \quad
    Meng Pang$^{4}$ \quad
    Yang Lu$^{1}$\thanks{Corresponding Author: Yang Lu (luyang@xmu.edu.cn)} \quad 
    Yan Yan$^{1}$ \quad
    Hanzi Wang$^{1}$ \\
    {$^1$Xiamen University} \quad
    {$^2$University of Cambridge}\quad
    {$^3$Zhejiang University} \quad
   {$^4$Nanchang University} \\
    {\tt\small yanshanshan@stu.xmu.edu.cn, zexi.li@zju.edu.cn, chao.wu@zju.edu.cn, }  \\
    {\tt\small mengpang@ncu.edu.cn, luyang@xmu.edu.cn, yanyan@xmu.edu.cn, hanzi\_wang@163.com}
}

\begin{document}
\maketitle
\begin{abstract}

Data heterogeneity, stemming from local non-IID data and global long-tailed distributions, is a major challenge in federated learning (FL), leading to significant performance gaps compared to centralized learning. 
Previous research found that poor representations and biased classifiers are the main problems and proposed neural-collapse-inspired synthetic simplex ETF to help representations be closer to neural collapse optima. However, we find that the neural-collapse-inspired methods are not strong enough to reach neural collapse and still have huge gaps to centralized training. In this paper, we rethink this issue from a self-bootstrap perspective and propose FedYoYo (You Are Your Own Best Teacher), introducing Augmented Self-bootstrap Distillation (ASD) to improve representation learning by distilling knowledge between weakly and strongly augmented local samples, without needing extra datasets or models. We further introduce Distribution-aware Logit Adjustment (DLA) to balance the self-bootstrap process and correct biased feature representations. FedYoYo nearly eliminates the performance gap, achieving centralized-level performance even under mixed heterogeneity. It enhances local representation learning, reducing model drift and improving convergence, with feature prototypes closer to neural collapse optimality. Extensive experiments show FedYoYo achieves state-of-the-art results, even surpassing centralized logit adjustment methods by 5.4\% under global long-tailed settings. 
% The code is available at \href{https://anonymous.4open.science/r/FedYoYo-1F01}{https://anonymous.4open.science/r/FedYoYo-1F01}.
\end{abstract}

\vspace{-10px}
\section{Introduction}

\label{sec:intro}
\begin{figure}
  \centering
    \includegraphics[width=0.99\linewidth]{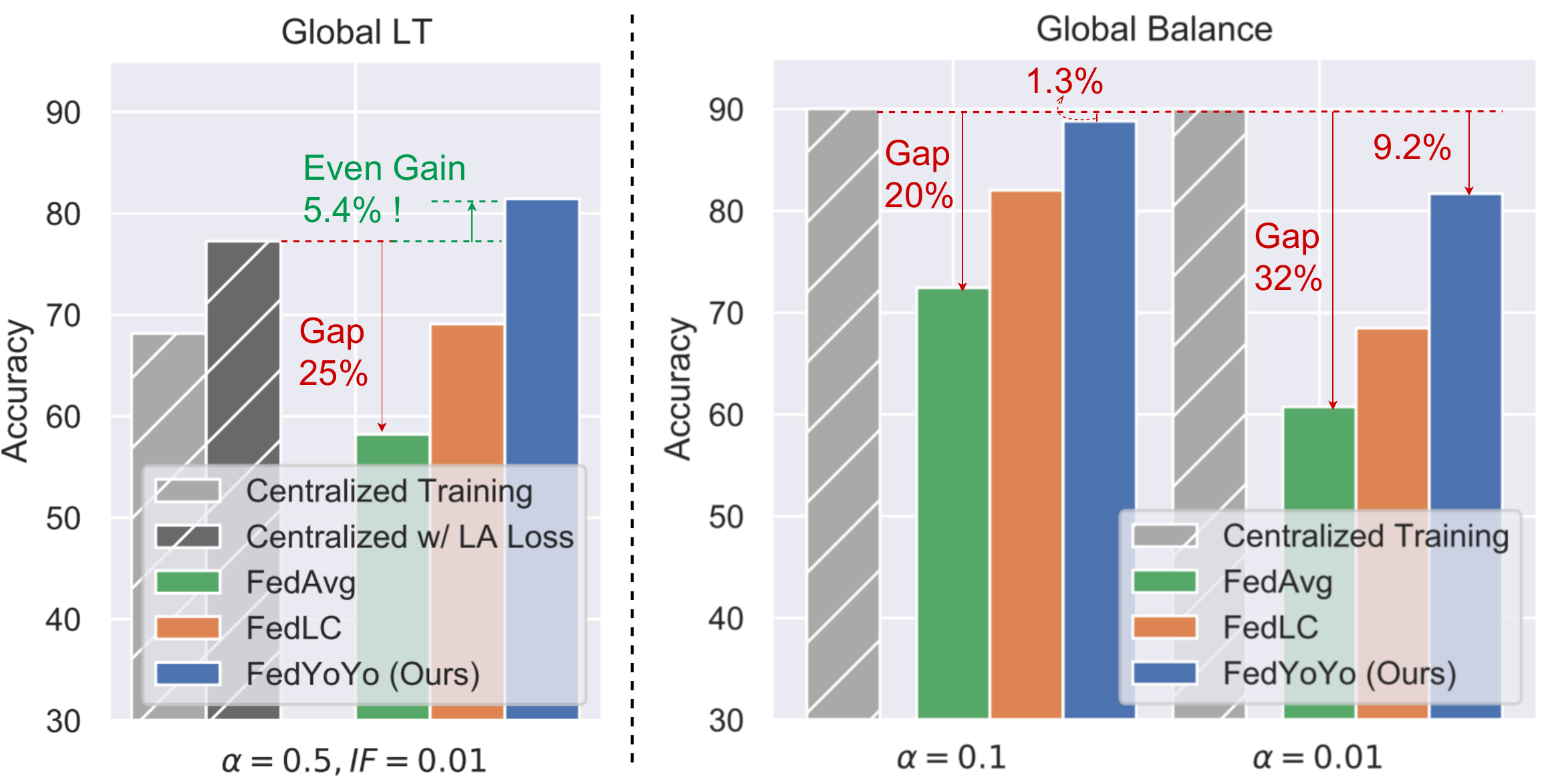}
  \caption{\textbf{Our method substantially closes the gap between centralized training and federated learning under heterogeneous data.} \textbf{Left:} non-IID data with global long-tailed (LT) distribution, where non-IID $\alpha=0.5$ and imbalance factor (IF) $= 0.01$. \textbf{Right:} non-IID data with global balanced distribution (vanilla non-IID setting in the literature: the smaller $\alpha$, the more non-IID). Baselines: vanilla centralized training, LA loss~\cite{la_loss}, FedAvg~\cite{fedavg}, FedLC~\cite{fedLC}. The results show that our method has marginal gaps with centralized training under vanilla non-IID data, but it can even surpass the centralized counterparts when the overall data distribution is long-tailed.}
  \label{fig:first_fig}
  \vspace{-10px}
\end{figure}

Federated learning (FL)~\cite{fedavg,fedlaw,scaffold,fedprox} is a collaborative learning paradigm that builds machine learning models from distributed data sources without sharing the raw data, which is communication-efficient~\cite{fedavg} and privacy-preserving~\cite{ppfl_survey}. FL is promising in a wide range of scenarios, like medical imaging~\cite{fl_medical_image}, multi-media analysis~\cite{fl_mm}, the Internet of Things~\cite{fl_iot}, etc. One inherent and key challenge in FL is data heterogeneity,  a critical bottleneck that significantly impacts the performance of FL methods~\cite{fedETF,fedprox}.

In practice, the overall data distribution in an FL system is often long-tailed, meaning that data heterogeneity arises from both the local non-IID data and the global long-tailed~\cite{creff,fedic,lt_survey} data. This combination leads to a more severe form of heterogeneity, causing in sub-optimal local models and poor generalization of the aggregated global model. As a result, there is always a huge gap between federated methods and centralized training due to data heterogeneity, which becomes even more evident when clients' data are highly non-IID~\cite{fedprox,fedavg,fedETF,scaffold}.

\begin{figure*}[t]
  \centering
  \includegraphics[width=\linewidth]{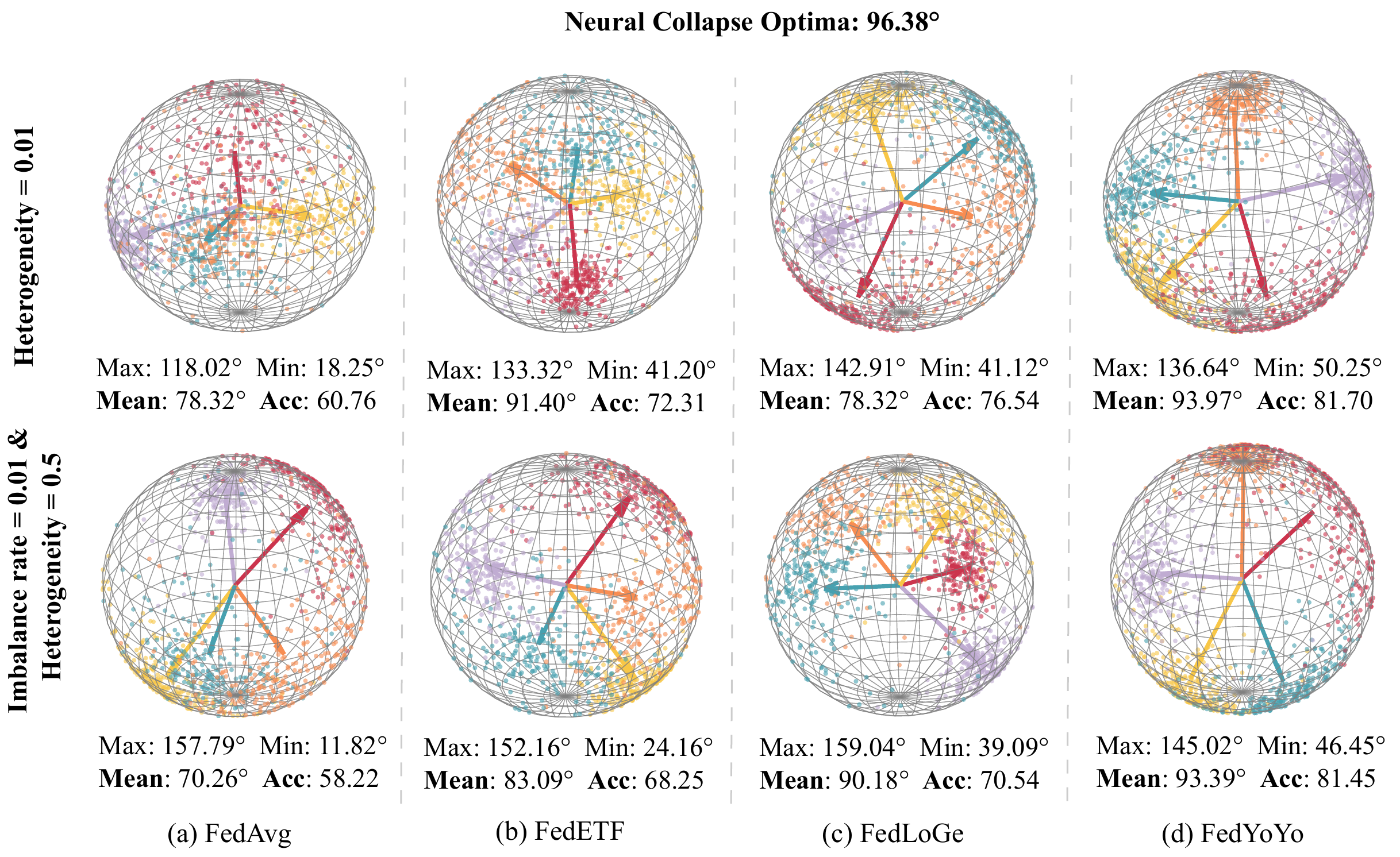}
  \caption{\textbf{Visualization of neural collapse degrees and accuracy for global models on CIFAR-10.} The optimal collapse angle (96.38°) follows Equiangular Tight Frame (ETF) theory, which defines ideal class prototype angles for maximal separation. Arrows indicate prototype representations, and colors denote different categories. \textbf{``Max'', ``Min'', and ``Mean''} represent the largest, smallest, and average angles between class prototypes, while \textbf{``Acc''} denotes global model accuracy. The top row shows vanilla data heterogeneity, and the bottom row includes long-tailed federated non-IID settings. (a) FedAvg, (b) FedETF, (c) FedLoGe, and (d) FedYoYo. Our FedYoYo method reaches better neural collapse conditions and achieves the best performance. \looseness=-1} 
  \label{fig:motivation}
  \vspace{-10px}

\end{figure*}

 Previous works tried to tackle non-IID data and improve the performances of FL by a large margin from FedAvg, but the centralized-federated gap is also dominant (\cref{fig:first_fig}). Recent studies ~\cite{ccvr,fedETF,fedloge,fedU2,feduv} have shown that data heterogeneity can lead to poor feature representations and biased classifiers, which might be the main cause of bad generalization performances. The poor representations in local models further exacerbates the feature misalignment between the global model and other client models. Therefore, the key challenge is to learn better and more unified feature representations across clients with non-IID data distributions. Existing approaches, such as FedETF~\cite{fedETF} and FedLoGe~\cite{fedloge}, attempt to mitigate the effects of heterogeneity by leveraging the theory of Neural Collapse ~\cite{RC,siamese} and constructing a synthetic simplex equiangular tight frame (ETF) with maximal pairwise angles~\cite{ETF}. Neural collapse is a deep learning phenomenon, which depicts an ideal representation and classifier structure under balanced and sufficient training.
Though these methods are inspired by neural collapse, we find they are not good enough to reach neural collapse optima under severe heterogeneity and global long-tailed distribution. An intuitive visualization is in \cref{fig:motivation}: the pair-wise prototype angles of FedETF and FedLoGe are far away from the theoretical neural collapse optima. 
 
Therefore, rethinking the representation learning strategy in non-IID federated learning is needed. We notice that self-supervised learning (SSL) has been widely validated in large-scale representation learning tasks, demonstrating the ability to capture more robust feature representations. This suggests that we can think of \emph{supervised} FL in an \emph{unsupervised} way to gain better representations, but more tailored designs are needed. The question is \emph{how to effectively use class/label distributions under data heterogeneity}. Existing SSL uses sample-wise contrastive or bootstrap methods and learns feature extractors instead of the whole model. But in supervised heterogeneous FL, class-biased local data and long-tailed global data will make features prioritize majority classes while neglecting minority ones, also the local model representations are not aligned, causing severe model drifts.\looseness=-1

In this paper, we find that self-bootstrap representation learning can release its full potential under the supervision of distribution-aware logit adjustment, and propose \textbf{FedYoYo}: \textbf{Yo}u Are \textbf{Yo}ur Own Best Teacher for local clients. 
Unlike previous SSL, we learn logits as representations instead of features, and logit adjustment can serve as distribution guidance to improve the representation of minor classes and align the representations across clients.
FedYoYo consists of two core components: Augmented Self-bootstrap Distillation (ASD) and Distribution-aware Logit Adjustment (DLA). 
ASD is inspired by BYOL~\cite{byol} in SSL, but tailored designs are made in FL. We learn logits instead of features and use one local model as the teacher itself instead of two online and target models in BYOL. We use the logits of a weakly augmented sample as the teacher to guide the learning of a strongly augmented sample. 
DLA is a tailored version of logit adjustment for heterogeneous FL; considering the potential long-tailed global distribution, we realize a tradeoff between both local and global distributions. 

FedYoYo realizes a matched co-design of self-bootstrap learning and logit adjustment, reaching near-centralized performances under non-IID FL. In scenarios where the global data distribution follows a long-tailed pattern, FedYoYo even surpasses centralized training. Moreover, in standard non-IID settings, it almost eliminates the performance gap between federated learning and centralized learning (\cref{fig:first_fig}). In \cref{fig:motivation}, it can be seen that our FedYoYo has better representations closer to neural collapse optimality and reaches higher generalization, compared with ETF-based methods.\looseness=-1

Our main contributions are summarized as follows: 
\begin{itemize} 
    {\item We propose FedYoYo, which addresses two challenges: data heterogeneity and the combined issue of global long-tailed and local non-IID data. It has two key components: Augmented Self-bootstrap Distillation and Distribution-Aware Logit Adjustment. \looseness=-1}

    {\item FedYoYo achieves a new level of performance in FL, it is comparable to centralized training. FedYoYo closes the centralized-federated gap from 20\% to 1.3\% in vanilla $\alpha=0.1$ data heterogeneity, and it even surpasses the centralized method by 5.4\% under global long-tailed distribution. \looseness=-1}

     \item We provide a new perspective on solving non-IID data and the caused poor representation issues in FL. We make FL more applicable and promising by reaching performances comparable to those of centralized. 
\end{itemize}

%-------------------------------------------------------------------------

\section{Proposed Method}
\begin{figure*}
  \centering
  \vspace{-0.3cm}
     \includegraphics[width=\linewidth]{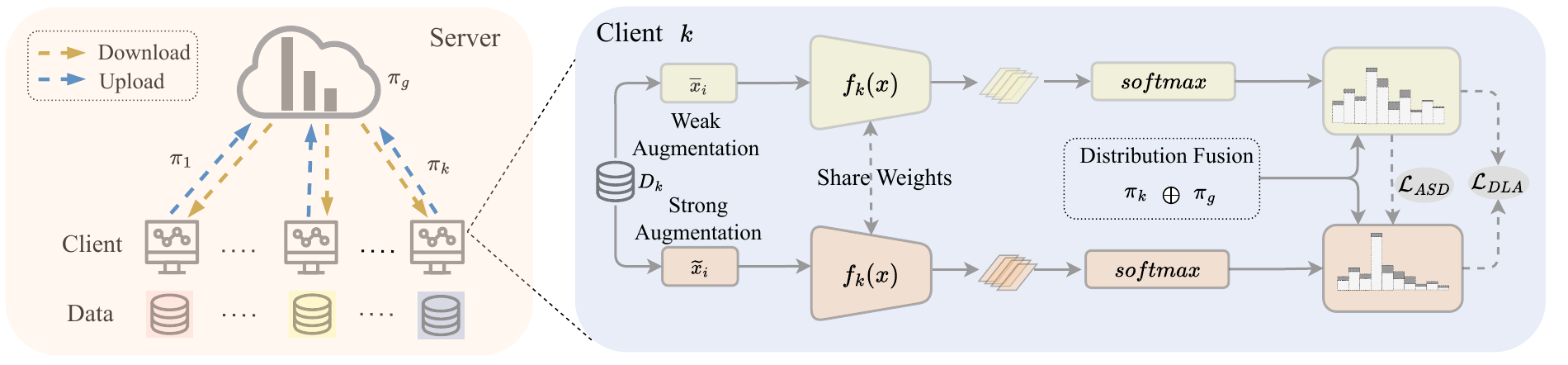}
     \vspace{-0.8cm}
    \caption{\textbf{Overview of our proposed FedYoYo framework.} On the server side, the estimated client distributions are aggregated to obtain an approximate global distribution. On the client side, client $k$ combines its local distribution $\pi_k$ with the global distribution $\pi_g$ using EMA updates for balance softmax. During local training, the model $f_k(x)$ processes weak and strong augmentations, using augmented self-bootstrap distillation loss ($\mathcal{L}_{ASD}$) and distribution-aware logit adjustment loss ($\mathcal{L}_{DLA}$) to enhance representation learning.}
    \vspace{-0.3cm}
    \label{fig:FedYoYo}
\end{figure*}

In this section, we introduce our proposed method FedYoYo, which focuses on enhancing feature representations and mitigating feature misalignment under data heterogeneity. The method consists of two key components: Augmented Self-bootstrap Distillation (ASD) and  Distribution-aware Logit Adjustment (DLA). ASD employs a self-bootstrap mechanism with logit adjustment as distribution guidance to enhance feature extraction, while DLA calibrates the classifier outputs using a fused global-local distribution to address client bias. Together, these components enable consistent and robust feature representation across clients, as illustrated in the overall framework ~\cref{fig:FedYoYo}.

\subsection{Preliminaries}
In the context of FL, consider a scenario with $K$ clients, where each client holds a non-IID local dataset $\mathcal{D}_k = \{(x_i, y_i) \mid 1 \leq i \leq n_k\}$, with $n_k$ representing the number of samples on client $k$, and $x_i$, $y_i$ denoting the input data and its corresponding label for each sample $i$. The model of client $k$ denotes $f_k(x)$. The entire training dataset $\mathcal{D} = \{\mathcal{D}_k\}_{k=1}^{K}$ combines the data from all clients. In vanilla FL, the overall distribution of $\mathcal{D}$ is class-balanced~\cite{fedprox,scaffold,fedETF}, while in our paper, we also consider a more realistic scenario where the global dataset exhibits a long-tailed distribution across the $C$ classes~\cite{creff}. In this global long-tailed distribution, we assume that classes are ordered by their sample frequencies such that if $i < j$, then $N^i \geq N^j$, where $N^i$ is the total number of samples for class $i$. For each class $c$, let $n_k^c$ represent the number of samples of class $c$ on client $k$, leading to the global sample count of class $c$ as $N^c = \sum_{k=1}^{K} n_k^c$. 

% 方法框架图

\subsection{Augmented Self-bootstrap Distillation}
We adopt a self-supervised learning approach similar to BYOL~\cite{byol}, applying different augmentations to local data to capture more robust feature representations. Then we introduce {\bf Augmented Self-bootstrap Distillation}, using the weakly augmented view as the teacher to provide guidance for the strongly augmented view, with the goal of enhancing feature representation quality. Unlike traditional self-distillation~\cite{self-kd2019}, which typically refers to knowledge transfer within a model such as from deeper to shallower layers, our approach distills knowledge across different views for the same model. Also, unlike BYOL, we learn logits instead of features and efficiently use one local model as the teacher itself instead of two online and target models in BYOL. Using logits as the representation can align the representation across clients since the logit space is the unified prediction space; also, we can calibrate the logits by considering both local and global distributions. 
Specifically, we employ Kullback-Leibler (KL) divergence to align the posterior distributions of the strong and weak augmentations. The distillation loss is then defined by minimizing the KL divergence as follows:

\begin{equation}
    \mathcal{L}_{ASD} = \frac{1}{n_k} \sum_{i=1}^{n_k} {KL} \big( p(\overline{x}_i) \,\big\|\, p(\widetilde{x}_i) \big),
    \label{equ:asd}
\end{equation}

where {$\overline{x}_i$} and {$\widetilde {x}_i$} represent the weak and strong augmentations of the same sample. Specifically, we leverage the local model to learn from two augmented instances generated through weak and strong augmentations. For weak augmentations, we apply techniques including RandomCrop, RandomHorizontalFlip, and RandomRotation. For strong augmentations, we follow \cite{autoaugment}. Notably, to avoid introducing noise, only correctly classified samples from the weak augmentation are used as the teacher to guide the learning of the corresponding student, rather than including all samples. Importantly, we adopt \cref{eq:BS-F} as the adjusted softmax $p(x)$ in \cref{equ:asd} to improve the performance of minority classes. \looseness=-1

The ASD technique not only captures richer features by leveraging a bootstrap style but also alleviates client drift by focusing on correctly classified samples. Thus, the local models and local data can be the best teachers for themselves in heterogeneous FL. This targeted guidance helps the local models converge more closely to optimal representations, bridging the performance gap between global and local models. Furthermore, by integrating an adjusted softmax, we effectively mitigate the biases caused by the long-tailed data, resulting in more stable and robust model performance across diverse clients.\looseness=-1

\subsection{Distribution-aware Logit Adjustment}
As mentioned, data heterogeneity often leads to biases in both feature representation and the classifier itself. Such biased classifiers may undermine the effectiveness of representation learning, thereby negatively affecting the local self-bootstrap distillation process. Therefore, it is essential to calibrate the classifier on each client during training to guarantee consistent outputs across all clients.
To achieve this, we adopt a balanced softmax for the model output, and the resulting probability distribution is:
\begin{equation}
    p(x)=\frac{n^y\exp\left(f(x,y)/T\right)}{\sum_{y^{\prime}=1}^Cn^{y^{\prime}}{\exp\left( {f(x,y^{\prime})}/T\right)}},
\end{equation}
where $n^y$ is the sample count of class $y$ and $T$ is the temperature coefficient (set to 1.5 in experiments) that controls the smoothness of the output. 

In federated long-tailed classification, global data imbalance exacerbates client-side heterogeneity, making sample count an unreliable measure of class representation. Minority classes can occupy large feature space regions despite few samples, while majority classes may show limited spread due to overlapping features. This reliance on sample count leads to biased class representation and degrades global model performance.

To address this, we propose a global-distribution-aware weight allocation strategy using the Pearson Correlation Coefficient to analyze local sample relationships for more adaptive class weighting. Inspired by previous work~\cite{area}, this method derives a local weight distribution that better reflects data variability. For class \( c \), we first compute the correlation matrix \( R_{cb} \) within each batch \( b \), which measures the pairwise relationships among all samples in the batch. The matrix \( R_{cb} \) is defined as:
\begin{equation}
\label{eq:cor_matrix_batch}
R_{cb} = \left[ R_{cb}^{i,j} \right]_{i,j \in b_k^c}, 
\end{equation}
where
\begin{equation}
R_{cb}^{i,j} = \frac{\left(h_i - \mu_{c}\right) \left(h_j - \mu_{c}\right)^\mathrm{T}}
{\left\| h_i - \mu_{c} \right\|_2 \cdot \left\| h_j - \mu_{c} \right\|_2}.
\end{equation}

where \( b_k^c \) denotes the set of samples from class $c$ in a single batch, $h_i$ and $h_j$ represent the feature vectors of samples $i$ and $j$, $\mu_{c}$  is the mean feature vector of class $c$ in the local data, i.e., the class prototype. Next, we compute the effective prior distribution iteratively across batches during training. The effective prior distribution $\pi_k^c$ is computed as:\looseness=-1
\begin{equation}
\pi_k^c = \left\{\sum_{b=1}^B \frac{1}{a_{cb}R_{cb}a_{cb}^\mathrm{T}}\right\},
\end{equation}
where $B$ denotes the total number of training batches, and $a_{cb} = \left(\frac{1}{|b_k^c|},\frac{1}{|b_k^c|},\dots,\frac{1}{|b_k^c|}\right) \in \mathbb{R}^{1 \times |b_k^c|}$. Finally, we obtain the local prior distribution as $\pi_k = \{\pi_{k}^1, \pi_{k}^2, \ldots, \pi_{k}^c\}$. 

Additionally, for data heterogeneity, we directly perform logit adjustment using the locally estimated distribution $ \pi_k$. However, in the case of federated long-tailed learning, the influence of the global distribution must also be considered. To address this, we approximate the global distribution $\pi_g $ by aggregating client-side $ \pi_k$ at the server by FedAvg, enabling more balanced model training. The final fused distribution $\pi_{mix}$ is defined as:\looseness=-1

\begin{equation}
    \label{eq:fusion}
    \pi_{mix} \leftarrow (1 - \gamma) \cdot \pi_{g} + \gamma \cdot \pi_k, 
\end{equation}
where $\gamma$ controls the degree of integration, which is discussed in \cref{ablation}.\looseness=-1

Then the distribution-fused balanced softmax is expressed as:
\begin{equation}
    p(x)=\frac{\pi_{mix}^y\exp\left(f(x,y)/T\right)}{\sum_{y^{\prime}=1}^C\pi_{mix}^{y^{\prime}}{\exp\left( f(x,y^{\prime})/T\right)}}.
    \label{eq:BS-F}
\end{equation}
Thus, we propose a distribution-aware logit adjustment loss function, denoted as DLA. The loss function is formulated as follows:
\begin{equation}
   \mathcal{L}_{DLA} = -\frac{1}{2{n_k}}\sum_{i=1}^{2n_k} \log \left( p(\hat x_i) \right),
\end{equation}
where {$\hat{x}_i$} represents the $i$-th augmented sample (both weakly and strongly augmented), with $2n_k$ accounting for the two types of augmented samples. Since self-bootstrap distillation involves two augmented views, we apply $\mathcal{L}_{DLA}$ to both views to ensure balanced loss contributions.

\subsection{Training}

Our method's local loss function combines two components: distillation loss $\mathcal{L}_{ASD}$ and classifier balancing loss $\mathcal{L}_{DLA}$. The $\mathcal{L}_{ASD}$ loss, as defined earlier, improves local representation learning. To mitigate classifier bias and its negative impact on feature representation learning, we use the $\mathcal{L}_{DLA}$ loss to adjust the local classifier. During client-side training, $\mathcal{L}_{DLA}$ is applied to learn from hard labels (i.e., the one-hot label), with both augmented batches processed simultaneously by the model.
Thus, the total loss function is defined as: \looseness=-1
\begin{equation}
    \mathcal{L}_{all} = \mathcal{L}_{DLA} + \lambda \mathcal{L}_{ASD},
\end{equation}
where $\lambda$ is a hyperparameter that controls the weight of the distillation loss. The impact of this hyperparameter is discussed in \cref{ablation}. 
For the server aggregation, we conduct FedAvg to local models.

\subsection{Privacy Discussion} 
Since local distribution estimation occurs entirely client-side, no privacy risk arises in the non-IID scenario. However, privacy concerns may emerge when incorporating global distribution information in federated long-tailed learning—a common challenge in federated learning, not specific to our method, as previous approaches such as Fed-Grab~\cite{fedgrab} and FedLoGe~\cite{fedloge} have also utilized local data distributions. If privacy protection is required, differential privacy (DP)~\cite{DP} can be applied by adding noise to uploaded distributions. A comprehensive discussion on federated learning privacy is beyond this work's scope; thus, we briefly address it here.

\section{Experiments}
\subsection{Experimental Setup}
\noindent {\bf Datasets and models.} We first evaluate our model on CIFAR-10/100, where the heterogeneity of the client data is controlled using the concentration parameter $\alpha$ of the Dirichlet distribution (the smaller $\alpha$, the more heterogeneous data). To further verify the robustness of our method under the more heterogeneous scenarios induced by real-world long-tailed distributions, we conduct experiments on several standard long-tailed datasets: CIFAR-10/100-LT, SVHN-LT, and ImageNet-LT. The imbalance factor (IF) is used to control the degree of imbalance. ImageNet-LT is a long-tailed version of ImageNet, with the largest and smallest categories containing 1,280 and 5 images. For CIFAR-10/100-LT and SVHN-LT, we utilize the ResNet-8 model, while for ImageNet-LT, we employ the ResNet-50 model. The detailed data distribution of CIFAR-10/100-LT is provided in Appendix.~\ref{data_show}.

\noindent {\bf Implementation of baseline methods.} We select three categories of state-of-the-art (SOTA) baseline methods for comparison: (1) Heterogeneity-oriented methods (FedProx~\cite{fedprox}, FedETF~\cite{fedETF}, FedLC~\cite{fedLC}, and CCVR~\cite{ccvr}) and federated long-tailed methods (CReFF~\cite{creff}, Fed-Grab~\cite{fedgrab}, BalanceFL~\cite{balancefl}, FedIC~\cite{fedic}, RUCR~\cite{rucr}, and FedLoGe~\cite{fedloge}); (2) Federated distillation methods, including FedDF~\cite{feddf}, FedFTG~\cite{fedftg}, FedGen~\cite{fedgen}, and DaFKD~\cite{dafkd}; (3) Long-tailed methods like $\tau$-norm~\cite{norm/lws}, AREA~\cite{area}, and LWS~\cite{norm/lws}. We also present the performance of centralized learning (CL) methods under both heterogeneous and long-tailed settings as an oracle upper bound.

\noindent {\bf Federated environment and local training.} We follow the experimental setup in previous federated long-tailed literature~\cite{creff}. We train for 300 rounds to reach sufficient convergence. All models are implemented in PyTorch and trained on NVIDIA GeForce 3090 GPUs. 

\subsection{Comparison with State-of-the-art Methods}

\begin{table}[t]
\begin{center}
\setlength\tabcolsep{10.0pt}
\caption{\textbf{Top-1 test accuracy ({\%}) of FedYoYo and FL methods with diffirent {$\alpha$}}. We compare our method on the vanilla non-IID setting without long-tailed distribution on  CIFAR-10/100. \looseness=-1}
\scalebox{0.80}{
\begin{tabular}{l|cc|cc} 
\toprule
\multirow{2}{*}{Method} & \multicolumn{2}{c|}{CIFAR-10} & \multicolumn{2}{c}{CIFAR-100}   \\
% \cline{2-10}
\cmidrule(lr){2-5}
& {$\alpha$=0.01} & {$\alpha$=0.1} & {$\alpha$=0.01} & {$\alpha$=0.1}         \\
% \hline
\midrule
\textbf{Centralized}  &  \multicolumn{2}{c|}{\textbf{90.03}} &  \multicolumn{2}{c}{\textbf{68.23}} \\
% \hline
\midrule
FedAvg  \cite{fedavg}    & 60.76	& 72.46	& 52.46	& 58.18 \\
FedProx \cite{fedprox}	& 55.22	& 70.71	& 52.91	& 60.37 \\
CCVR \cite{ccvr}	    & 63.25	& 74.18	& 56.95	& 61.82 \\
FedDF \cite{feddf} & 60.83 & 79.48&44.49 &51.69 \\
FedGen \cite{fedgen}& 52.00& 71.82&37.72 &43.44\\
FedFTG \cite{fedftg}& 51.54&79.07 &47.87 &53.93\\
DaFKD \cite{dafkd}& 59.28&81.67 & 47.15&52.17\\
\midrule
FedLC 	\cite{fedLC}    & 68.49	& 82.03	& 58.11	& 60.44 \\
FedETF 	\cite{fedETF}    & 72.31	& 83.10	& 58.18	& 61.87 \\
% \rowcolor{darkgreen!10}  % 
\textbf{FedYoYo} & \textbf{81.70}	& \textbf{88.82}	& \textbf{64.49}	& \textbf{67.59} \\
\bottomrule
\end{tabular}}
\label{tab:only-noniid}
\end{center}
\vspace{-0.5cm}
\end{table}

% 大表
\begin{table*}[t]
\begin{center}
% \vspace{-2mm}
% \vspace{-0.2cm}
\caption{\textbf{Top-1 test accuracy (\%) of FedYoYo and SOTA methods on CIFAR-10/100-LT, SVHN-LT with different IFs and {$\alpha=0.5$}.} The best performance is in bold, and the second is underlined. ~\textcolor{darkred}{Red} text indicates improved performance compared to the centralized learning with LA loss. ~\textcolor{darkgreen}{\bf $\uparrow$} indicates improved accuracy compared with the underlined (best baseline in each setting).  }
\vspace{-0.1cm}
\setlength\tabcolsep{6pt}
% \hspace{-0.2cm}
\scalebox{0.65}{
\begin{tabular}{lccccccccc} 
%\cline{1-13}
\toprule

\multirow{2}{*}{Method} & \multicolumn{3}{c}{CIFAR-10-LT} & \multicolumn{3}{c}{CIFAR-100-LT} &\multicolumn{3}{c}{SVHN-LT}  \\
% \cline{2-10}
\cmidrule(lr){2-10}

& IF=100 & IF=50 & IF=10 & IF=100 & IF=50 & IF=10  & IF=100 & IF=50 & IF=10         \\
\hline
\midrule
\multicolumn{10}{l}{\bf Centralized Methods}\\
\hline
\midrule
Centralized & 68.17 & 76.15 & 80.32 & 38.61 & 40.30 & 50.34 & 84.86 & 86.55 & 90.14  \\
Centralized w/ LA Loss \cite{la_loss} & \textbf{77.27} & \textbf{80.23} & \textbf{86.15} & \textbf{41.57} & \textbf{44.02} & \textbf{56.06} & \textbf{88.17} & \textbf{90.23} & \textbf{93.45}  \\
\hline
\midrule
\multicolumn{10}{l}{\bf Heterogeneity-oriented FL methods}\\
\hline
\midrule
FedAvg \cite{fedavg}     & 58.22	& 63.47	& 77.95	& 31.64	& 36.14	& 46.84	& 81.31	& 84.40	& 87.31 \\
FedProx \cite{fedprox}	& 55.68	& 60.72	& 76.64	& 32.73	& 35.82	& 45.77	& 82.04	& 86.33	& 90.13 \\
CCVR \cite{ccvr}	& 68.13	& 72.98	& 81.56	& 34.73	& 37.68	& 48.92	& 82.34	& 86.48	& 91.26 \\
FedLC 	\cite{fedLC}    & 69.02	& 71.93	& 79.09	& 31.75	& 38.36	& 47.54	& 81.17	& 85.23	& 87.43 \\
FedETF 	\cite{fedETF}    & 68.25	& 72.35	& 81.60	& 33.19	& 37.86	& 48.71	& 83.02	& 87.07	& 90.86 \\
\hline
\midrule
\multicolumn{10}{l}{\bf Imbalance-oriented FL methods}\\
\hline
\midrule
{$\tau$}-norm \cite{norm/lws}	& 40.70	& 41.31	& 51.66	& 19.29	& 20.82	& 32.51	& 70.65	& 74.63	& 78.58 \\
LWS  \cite{norm/lws}	& 37.46	& 39.82	& 49.30	& 18.18	& 20.10	& 33.81	& 71.18	& 73.25	& 78.15 \\
AREA  \cite{area}	& 64.33	& 65.16	& 78.97	& 36.34	& 37.83	& 48.59	& 82.87	& 85.24	& 90.35 \\
\hline
\midrule
\multicolumn{10}{l}{\bf Federated long-tailed methods}\\
\hline
\midrule
BalanceFL \cite{balancefl}	& 49.34	& 54.17	& 72.03	& 26.03	& 29.28	& 40.16	& 75.41	& 79.13	& 85.16 \\
CReFF \cite{creff}	& 70.51	& 73.60	& 79.85	& 32.90	& 34.66	& 43.42	& 85.47	& 87.64	& 91.16  \\
FedIC \cite{fedic}	& 66.49	& 67.55	& 71.21	& 33.67	& 36.74	& 41.93	& 84.94	& 86.82	& 90.53 \\
Fed-Grab \cite{fedgrab}	& \underline{70.63}	& \underline{75.44}	& \underline{85.21}	& 33.53	& 44.01	& 55.87	& \underline{90.39}	& \underline{91.11}	& \underline{94.56} \\
RUCR \cite{rucr}	& 55.32	& 60.24	& 75.56	& 27.61	& 33.81	& 41.29	& 73.45	& 82.20	& 87.23 \\
FedLoGe \cite{fedloge}	& 70.54	& 74.85	& 84.54	& \underline{42.63}	& \underline{47.66}	&\underline{58.36}	& 85.05	& 87.06	& 89.68 \\
\midrule

{\bf FedYoYo}	& {\bf 81.45 }(\textcolor{darkred}{\bf +4.18})	& {\bf 83.85 }(\textcolor{darkred}{\bf +3.62}) & {\bf 87.94 }(\textcolor{darkred}{\bf +1.79})	& {\bf 46.13 }(\textcolor{darkred}{\bf +4.56})& {\bf 50.83 }(\textcolor{darkred}{\bf +6.81}) & {\bf 60.16 }(\textcolor{darkred}{\bf +4.10})& {\bf 91.73 }(\textcolor{darkred}{\bf +3.56})	& {\bf 92.38 }(\textcolor{darkred}{\bf +2.15})	& {\bf 95.10 }(\textcolor{darkred}{\bf +1.65}) \\
& \textcolor{darkgreen}{\bf $\uparrow$ 10.82} & \textcolor{darkgreen}{\bf $\uparrow$ 8.41} & \textcolor{darkgreen}{\bf $\uparrow$ 2.73} & \textcolor{darkgreen}{\bf $\uparrow$ 3.50}	& \textcolor{darkgreen}{\bf $\uparrow$ 3.17} & \textcolor{darkgreen}{\bf $\uparrow$ 1.80} & \textcolor{darkgreen}{\bf $\uparrow$ 1.34} & \textcolor{darkgreen}{\bf $\uparrow$ 1.27} & \textcolor{darkgreen}{\bf $\uparrow$ 0.54} 	 \\
\bottomrule
\end{tabular}}
\label{tab:main-table}
\end{center}
\vspace{-0.5cm}
\end{table*}

\noindent {\bf Results on vanilla non-IID settings.}  
All results are reported in \cref{tab:only-noniid}. Our FedYoYo achieves the best performance under different $\alpha$. It is notable that our method surpasses FedETF and FedLC in extreme non-IID settings. FedETF is the state-of-the-art method using neural-collapse-inspired classifiers. The results show our method has better generalization than FedETF and \cref{fig:motivation} shows that our method can realize better neural collapse optimality than FedETF. FedLC incorporates logit adjustment in FL, but our co-design of self-bootstrap and logit adjustment can reach better performances than logit adjustment solely (i.e., FedLC). 
Furthermore, our approach successfully reduces the performance gap with the centralized baseline, demonstrating its effectiveness in mitigating the challenges posed by non-IID data distribution.\looseness=-1

\begin{table}[t]
\begin{center}
\caption{\textbf{Top-1 test accuracy ({\%})  of FedYoYo and SOTA methods on ImageNet-LT with  {$\alpha=0.1$}. }}
\vspace{-0.2cm}
\setlength\tabcolsep{10.0pt}
\scalebox{0.80}{
\begin{tabular}{lcccc} 
\toprule
\multirow{2}{*}{Method} & \multicolumn{4}{c}{ImageNet-LT}   \\
% \cline{2-10}
\cmidrule(lr){2-5}
& Many & Medium & Few & All         \\
\midrule
FedAvg  \cite{fedavg}    & 34.92	& 19.18	& 7.41	& 23.85 \\
FedProx \cite{fedprox}	& 34.25	& 17.06	& 6.73	& 22.57 \\
CCVR \cite{ccvr}	    & 36.72	& 20.24	& 9.26	& 25.49 \\
FedLC 	\cite{fedLC}    & 36.03	& 21.14	& 6.57	& 23.23 \\
FedETF 	\cite{fedETF}    & 35.94	& 19.91	& 7.07	& 23.97 \\
\midrule
{$\tau$}-norm \cite{norm/lws}	& 30.81	& 14.57	& 5.22	& 19.58 \\
LWS \cite{norm/lws}	& 37.23	& 23.4	& 7.50	& 25.37 \\
AREA  \cite{area}	& 39.83	& 23.51	& 8.53	& 26.03 \\
\midrule
BalanceFL \cite{balancefl}	& 30.63	& 19.26	& 7.20	& 21.87 \\
CReFF \cite{creff}	& 37.61	& 21.48	& 10.02	& 26.91  \\
FedIC \cite{fedic}	& 36.22	& 20.5	& 9.76	& 25.71 \\
Fed-Grab \cite{fedgrab} & 41.16	& 24.42	& 14.29	& 30.56 \\
RUCR \cite{rucr}	 & 30.47	& 15.77	& 5.22	& 20.06 \\
FedLoGe \cite{fedloge} & 40.77	& 24.16	& 13.27	& 29.73 \\
\midrule
% \rowcolor{darkgreen!10}  
{\bf FedYoYo}	& {\bf 42.05}	& {\bf 25.78} & {\bf 15.44} & {\bf 31.41}	  \\
\bottomrule
\end{tabular}}
\label{tab:imagenet-lt}
\end{center}
\vspace{-0.5cm}
\end{table}

\noindent {\bf Results on global long-tailed and non-IID settings.} \textit{\textbf{CIFAR-10/100-LT and SVHN-LT:}} As summarized in \cref{tab:main-table}, our method achieves the highest test accuracy across all datasets with varying imbalance factors (IFs). Heterogeneity-oriented methods generally perform similarly to FedAvg, as they focus on data heterogeneity but overlook global class imbalance. Imbalance-oriented methods like AREA perform better than FedAvg in certain cases but still lag behind our approach, likely because they primarily address imbalance without accounting for inter-client data heterogeneity. Notably, our method consistently shows significant performance gains over centralized learning with LA loss. 
\textit{\textbf{ImageNet-LT:}} We further validate our method on the more challenging ImageNet-LT dataset. In \cref{tab:imagenet-lt}, we report accuracies for three class groups: many-shot (over 100 samples per class), medium-shot (20-100 samples), and few-shot (fewer than 20 samples). Our method consistently outperforms others in all groups, particularly in few-shot classes, where it achieves 15.44\% accuracy—an 8.04\% improvement over the baseline. This showcases the effectiveness of our approach in enhancing few-shot class performance while maintaining strong accuracy in many-shot classes.

\subsection{Analysis of FedYoYo}

\begin{figure}[t]
  \centering
    \includegraphics[width=\linewidth]{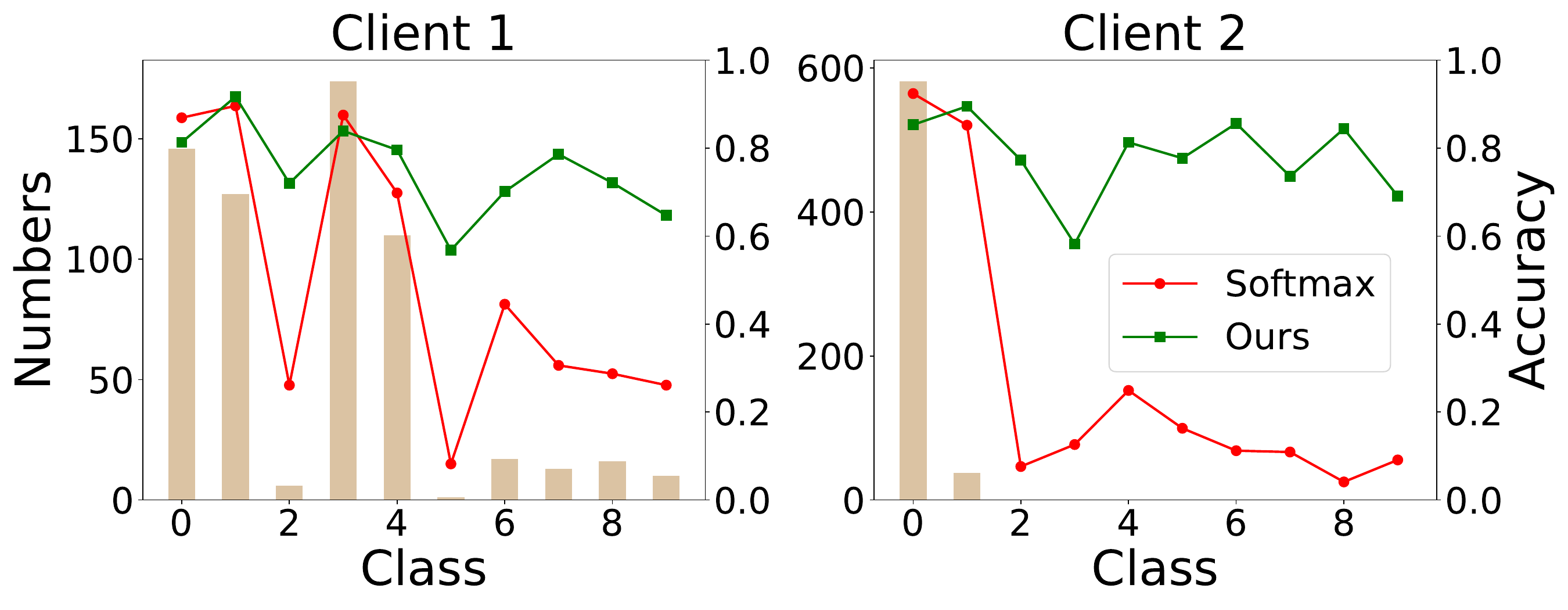}
    \vspace{-0.6cm}
\caption{\textbf{Comparison of per-class local accuracy after receiving the global model and performing local updates.} Bars represent the local data distribution, while lines indicate accuracy. ``Softmax'' means FedAvg with vanilla softmax.}
\label{fig:local-acc}
\vspace{-0.2cm}
\end{figure}
\noindent{\bf Global-to-local model gap reduction.} We visualize the average gains from local training and global aggregation in Appendix.~\ref{fedyoyo_analsys}. Our method significantly reduces the gap between global and local models compared to FedAvg, enhancing the gains from global aggregation. This demonstrates that our approach benefits the global model, enabling faster adaptation to local data, even under the challenging non-IID conditions caused by long-tailed distributions. As local models converge toward optimal consistency, the aggregated model also achieves superior performance. Furthermore, as shown in \cref{fig:local-acc}, our method surpasses FedAvg in creating more balanced across-client performance. Overall, our approach exhibits strong adaptability in heterogeneous and long-tailed scenarios, resulting in notable improvements for both global and local models.\looseness=-1

\noindent{\bf Effectiveness of estimated global distribution.} Our estimated class distribution closely aligns with the oracle distribution. To further verify this, we tracked the $\ell_2$ distance between the estimated and original data distributions across training epochs in Appendix.~\ref{fedyoyo_analsys}.

% TSNE
\begin{figure}[t]
  \centering
    \includegraphics[width=\linewidth]{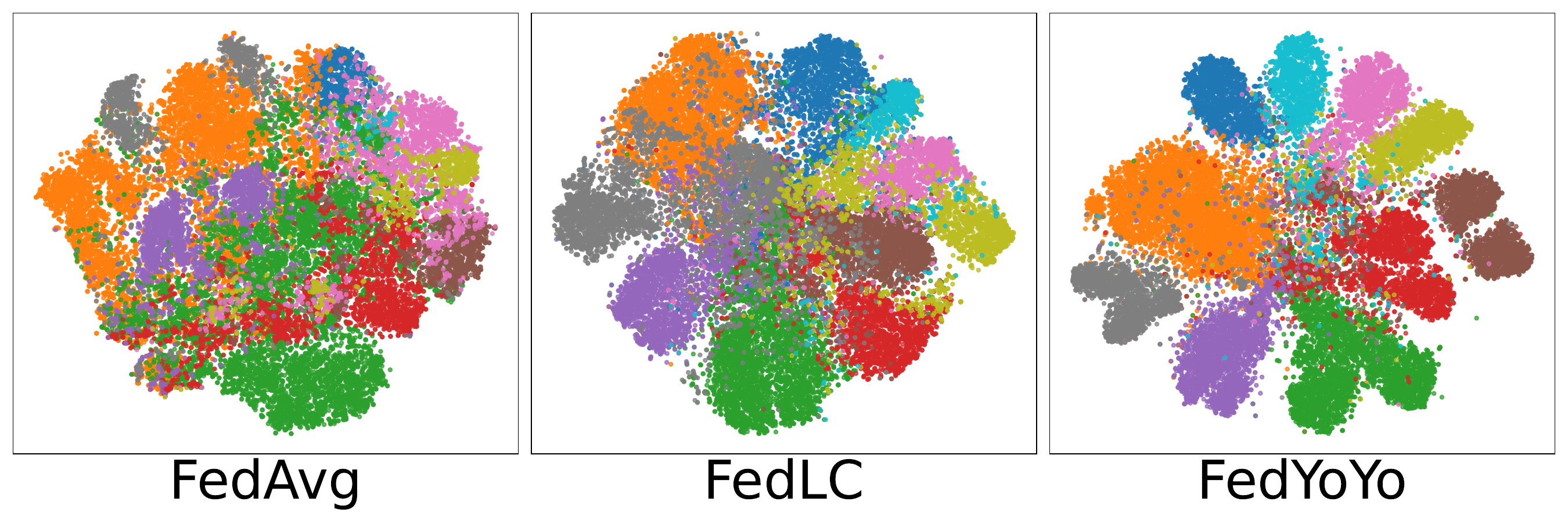}
    \vspace{-0.6cm}
    \caption{\textbf{The t-SNE visualization of feature spaces of FedAvg~\cite{fedavg}, FedLC~\cite{fedLC}, and our FedYoYo.}}
  \label{fig:tsne}
  \vspace{-0.4cm}
\end{figure}
\noindent {\bf Feature level analysis.} In \cref{fig:tsne}, we present a visualization of the feature representations extracted by the global model. Compared to other methods, our approach achieves more compact intra-class features and better inter-class separability. Additionally, we compare the similarity between features of the global model and those of the local models in Appendix.~\ref{fedyoyo_analsys}. Results show that our method effectively reduces the feature discrepancy between local and global models, demonstrating its ability to alleviate client drift and improve local representation learning. \looseness=-1

\noindent{\bf Computation Cost.} We evaluated three methods using ResNet-18 on CIFAR-100-LT (IF=100) in terms of GFLOPs per round. Compared to FedAvg's 1003.72G and FedGrab's 2371.36G, our method, FedYoYo, achieves better performance with a more efficient computational cost of 1721.79G. \looseness=-1

\subsection{Ablation Study}\label{ablation}
\begin{figure}
  \centering
  \begin{subfigure}{0.49\linewidth}
    \includegraphics[width=\linewidth]{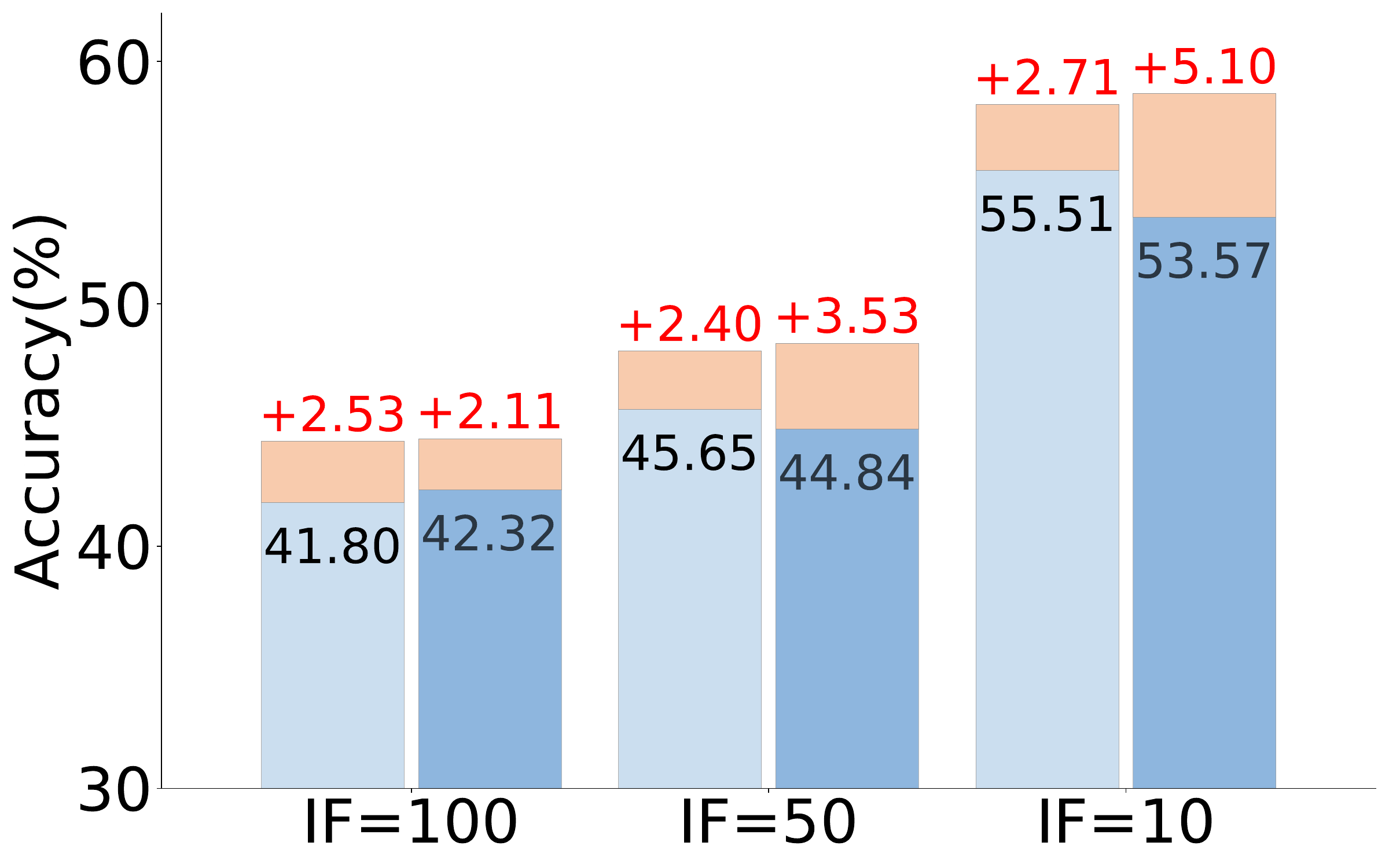}
    \caption{CIFAR-100-LT, {$\alpha$}=0.1.}
    \label{fig:eff_mix(noniid=0.1)}
  \end{subfigure}
  \hfill
  \begin{subfigure}{0.49\linewidth}
    \includegraphics[width=\linewidth]{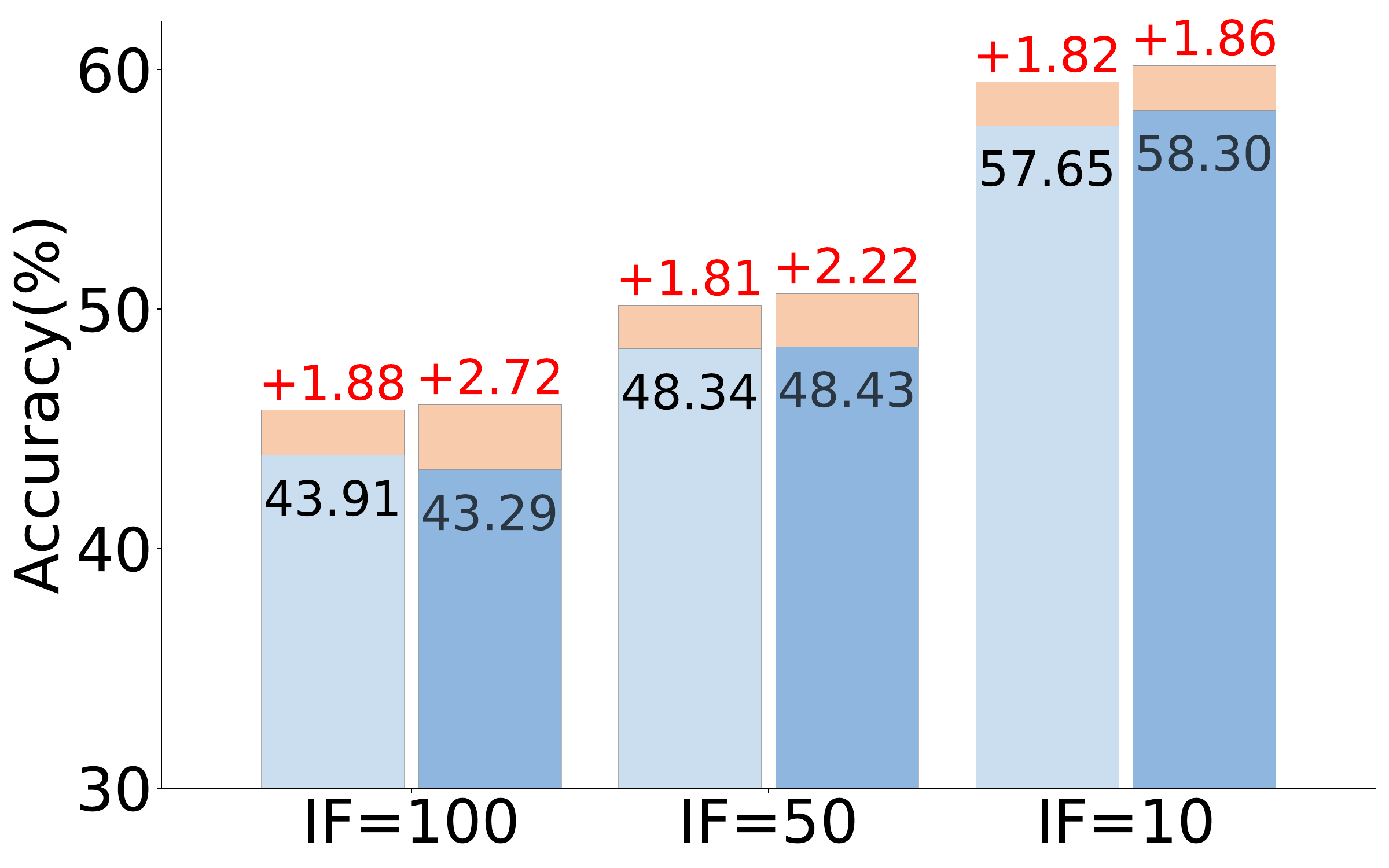}
    \caption{CIFAR-100-LT, {$\alpha$}=0.5.}
    \label{fig:eff_mix(noniid=0.5)}
  \end{subfigure}
  \vspace{-0.2cm}
  \caption{\textbf{Top-1 test accuracy ({\%}) of our method using different distribution estimation strategies on CIFAR-100-LT with varying $\alpha$.} Light blue bars show results using the ground-truth distribution, while dark blue bars represent the Pearson coefficient-based estimation. Red texts highlight the performance gains from the fused distribution strategy.}
  \label{fig:eff-mix}
  \vspace{-0.3cm}
\end{figure}

\noindent{\bf The necessity of distribution fusion.} We compare the performance of the real distribution and the estimated distribution and further evaluate the effect of applying a fusion strategy to each. As illustrated in \cref{fig:eff-mix}, after applying the fusion strategy, significant improvements are observed in both cases. Notably, under high heterogeneity settings (e.g., {$\alpha$=0.1}), the estimated distribution with the fusion strategy even outperforms the real distribution, demonstrating that the fusion approach not only mitigates the impact of data heterogeneity but also enhances model generalization. This confirms the value of fusing global and local distributions in strengthening model robustness.

\begin{table}[t]
\centering
\caption{\textbf{Ablation study of our method's key components on the CIFAR-100-LT dataset with IF $= 100$ and  {$\alpha=0.5$}.}}
\vspace{-0.2cm}
\setlength\tabcolsep{6.0pt}
\scalebox{0.86}{
\begin{tabular}{ccc|ccc|c} 
\toprule
 RandAug & ASD  & DLA & Many & Medium & Few & All \\
\midrule
 & & &      56.89	& 31.94	& 7.50	& 33.34 \\
 {\large \ding{51}} & & & 60.18	& 30.49	& 12.50	& 35.71 \\
     & & {\large \ding{51}}&  51.46	& 34.43	& 21.13	& 36.40 \\
   {\large \ding{51}} & {\large \ding{51}}& & {\bf 65.31}	& 42.71	& 13.93	& 41.99 \\
 {\large \ding{51}}& & {\large \ding{51}}&  56.03	& 42.51	& 27.33	& 41.69 \\
{\large \ding{51}}& {\large \ding{51}} &{\large \ding{51}} &  60.26	& {\bf 47.00}	& {\bf 28.63}	& {\bf 46.13} \\
\bottomrule
\end{tabular}}
\label{tab:ablation}
% \vspace{-0.2cm}
\end{table}

\begin{table}[t]
\centering
\caption{\textbf{Top-1 test accuracy ({\%}) accuracy comparison with different augmentation policys on CIFAR100-LT with IF $= 100$ and {$\alpha=0.5$}.}}
\vspace{-0.2cm}
\setlength\tabcolsep{7pt}
\scalebox{0.9}{
\begin{tabular}{ccc} 
\toprule
View1(Teacher) & View2(Student) & Accuracy \\
\midrule
Weak augmentation & Weak augmentation& 33.74 \\
Strong augmentation& Strong augmentation & 42.92 \\
Weak augmentation& Strong augmentation& 46.13 \\
Strong augmentation& Weak augmentation& 41.06 \\
\bottomrule
\end{tabular}}
\label{tab:weak-strong-aug}
\vspace{-0.4cm}
\end{table}
\noindent{\bf Ablation studies on all components of FedYoYo.} In \cref{tab:ablation}, we present a comprehensive ablation study on the CIFAR-100-LT dataset, evaluating key components: RandAug, ASD, and DLA. RandAug refers to the use of strong augmentation. A vanilla model without any components reaches an accuracy of 33.34\%, similar to FedAvg. When RandAug is combined with ASD, accuracy improves by 8.65\%, highlighting ASD's crucial role in enhancing feature learning and knowledge transfer. Notably, without DLA, many-shot classes see a significant performance boost, but gains for few-shot classes are minimal. When all components are combined, the overall accuracy improves by 12.79\%, with gains of 8.42\% for medium-shot and 21.13\% for few-shot classes. These findings show that DLA mitigates classifier bias and balances the distillation process. Additionally, ASD and DLA reinforce each other, working synergistically to deliver consistent performance improvements across head, medium, and tail classes.

\noindent {\bf Impact of data augmentation.} In this section, we apply various augmentations to the training samples to assess the effectiveness of ASD. As shown in \cref{tab:weak-strong-aug}, we compare four augmentation variants in FedYoYo, and the results indicate that the weak-strong self-bootstrap distillation outperforms other augmentations. Moreover, other augmentation methods were examined in Appendix.~\ref{more_ablation}.

\begin{figure}
  \centering
  \begin{subfigure}{0.49\linewidth}
    \includegraphics[width=\linewidth]{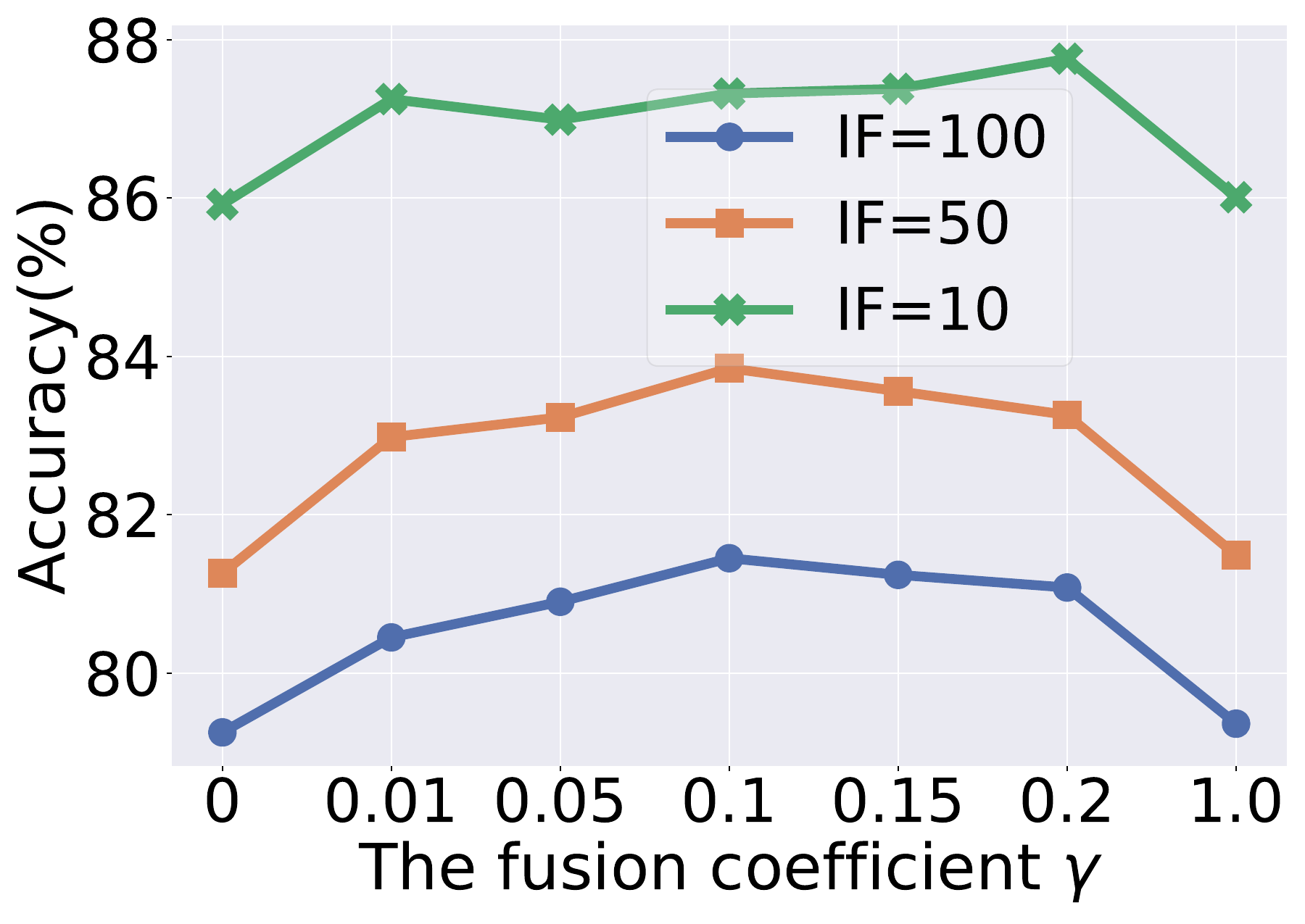}
    \caption{Influence of $\gamma$.}
    \label{fig:param_gamma}
  \end{subfigure}
  \hfill
  \begin{subfigure}{0.49\linewidth}
    \includegraphics[width=\linewidth]{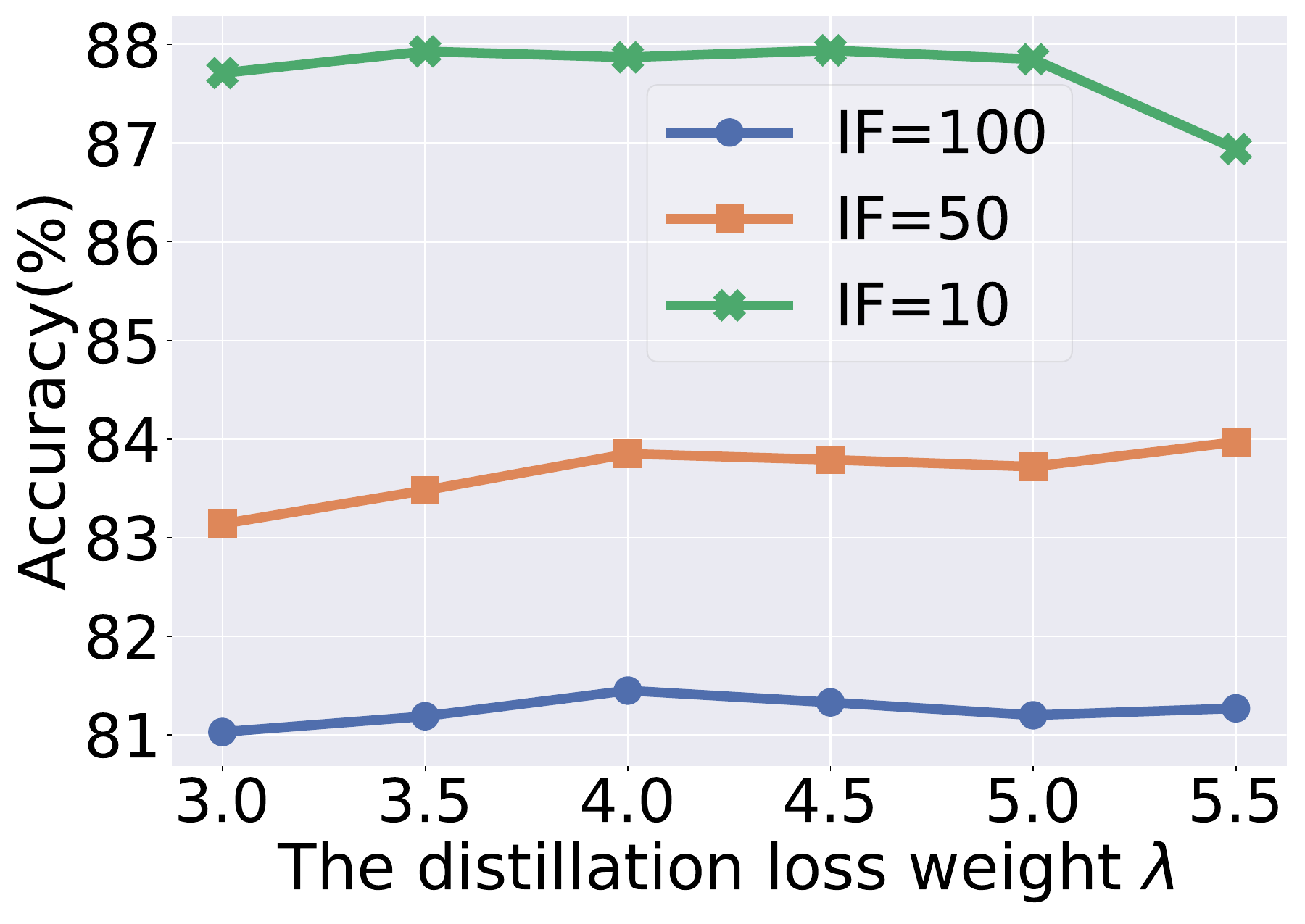}
    \caption{Influence of $\lambda$.}
    \label{fig:param_lambda}
  \end{subfigure}
   \vspace{-0.2cm}
\caption{\textbf{(a) Comparison of different fusion ratios $\gamma$.} We report the performance across various fusion ratios. \textbf{(b) Comparison of different values of $\lambda$.}\looseness=-1}
\label{fig:params}
\vspace{-0.6cm}
\end{figure}

\noindent {\bf Influence of fusion coefficient {$\gamma$}.} \cref{fig:param_gamma} shows the effect of the fusion coefficient $\gamma$ on model accuracy across different imbalance factors (IF). A larger $\gamma$ emphasizes the global distribution. Initially, increasing $\gamma$ improves accuracy, but excessive values (e.g., 1.0) cause a decline, indicating that relying solely on global or local distributions is suboptimal. Our approach balances global and local information, achieving better overall performance.

\noindent {\bf Influence of loss weight $\lambda$.} As shown in \cref{fig:param_lambda}, despite increasing $\lambda$ from 3.0 to 5.5, the model performance remains relatively stable with minor variations in accuracy across different values. This suggests that the model is not sensitive to the choice of $\lambda$, implying that consistent performance can be maintained even with different weights. Therefore, we recommend choosing $\lambda$ within this range, as it provides stable and reliable performance.

\section{Related Works}
\label{sec:Related Works}
\subsection{Data Heterogeneity in Federated Learning}

Our paper focuses on data heterogeneity in federated learning that includes local non-IID heterogeneity and global long-tailed data heterogeneity. Local data heterogeneity causes client drift, which is often mitigated by regularization methods like FedProx~\cite{fedprox}, SCAFFOLD~\cite{scaffold}, and Moon~\cite{moon}. To tackle classifier bias in non-IID settings, prior works explore strategies such as classifier retraining~\cite{ccvr}, prototype-based classifier rebalancing~\cite{2023tackling,2024fedtgp} and local calibration~\cite{fedLC}. However, Global long-tailed distributions further exacerbate cross-client heterogeneity, making adaptation more difficult. Recent methods focus on server-side calibration~\cite{creff, fedic} or local model adjustment using global distribution~\cite{fedgrab}, but their effectiveness remains limited, highlighting the need for more robust solutions.\looseness=-1

\subsection{Feature Representation Learning in FL}
Data heterogeneity in federated learning leads to poor feature representations and biased classifiers, causing misalignment between global and client models. Under global long-tailed distributions, classifier bias further worsens this misalignment. Anchor-based methods~\cite{fedfa,fedfm} attempt feature alignment, while neural-collapse-inspired approaches employ ETF-based methods~\cite{fedETF,fedloge} or feature regularization~\cite{feddecorr} to enhance generalization. Others use re-weighting in client aggregation~\cite{climb} to mitigate bias. However, these methods yield limited improvements in feature representation. Existing self-supervised federated methods typically utilize contrastive learning~\cite{simclr,SFL} or bootstrap methods~\cite{byol}, often overlooking minority classes and disproportionately emphasizing majority classes in supervised heterogeneous federated scenarios. In contrast, our method explicitly learns logits as representations and employs logit adjustment guided by distribution information. This strategy effectively enhances minority-class representation, aligns client features, and improves the global model's generalization and performance.

%-------------------------------------------------------------------------

\vspace{-0.3cm}
\section{Conclusions}
\vspace{-0.2cm}
We propose FedYoYo, a novel federated learning method addressing challenges posed by heterogeneous and long-tailed data distributions. FedYoYo integrates Augmented Self-bootstrap Distillation (ASD) and Distribution-aware Logit Adjustment (DLA). ASD employs weakly augmented samples as self-teachers to guide strongly augmented samples, enhancing local feature extraction under client data diversity. DLA leverages both local and global distributions to calibrate logits, providing effective guidance signals for representation learning. Extensive experiments on CIFAR-10-LT, CIFAR-100-LT, and ImageNet-LT demonstrate FedYoYo’s state-of-the-art performance, surpassing even centralized baselines in global long-tailed scenarios.\looseness=-1

\newpage
{
    \small
    \bibliographystyle{ieeenat_fullname}
    \bibliography{main}
}
\clearpage
\setcounter{page}{1}
\maketitlesupplementary
\appendix
\begin{figure}[t]
  \centering
  \begin{subfigure}{\linewidth}
    \includegraphics[width=\linewidth]{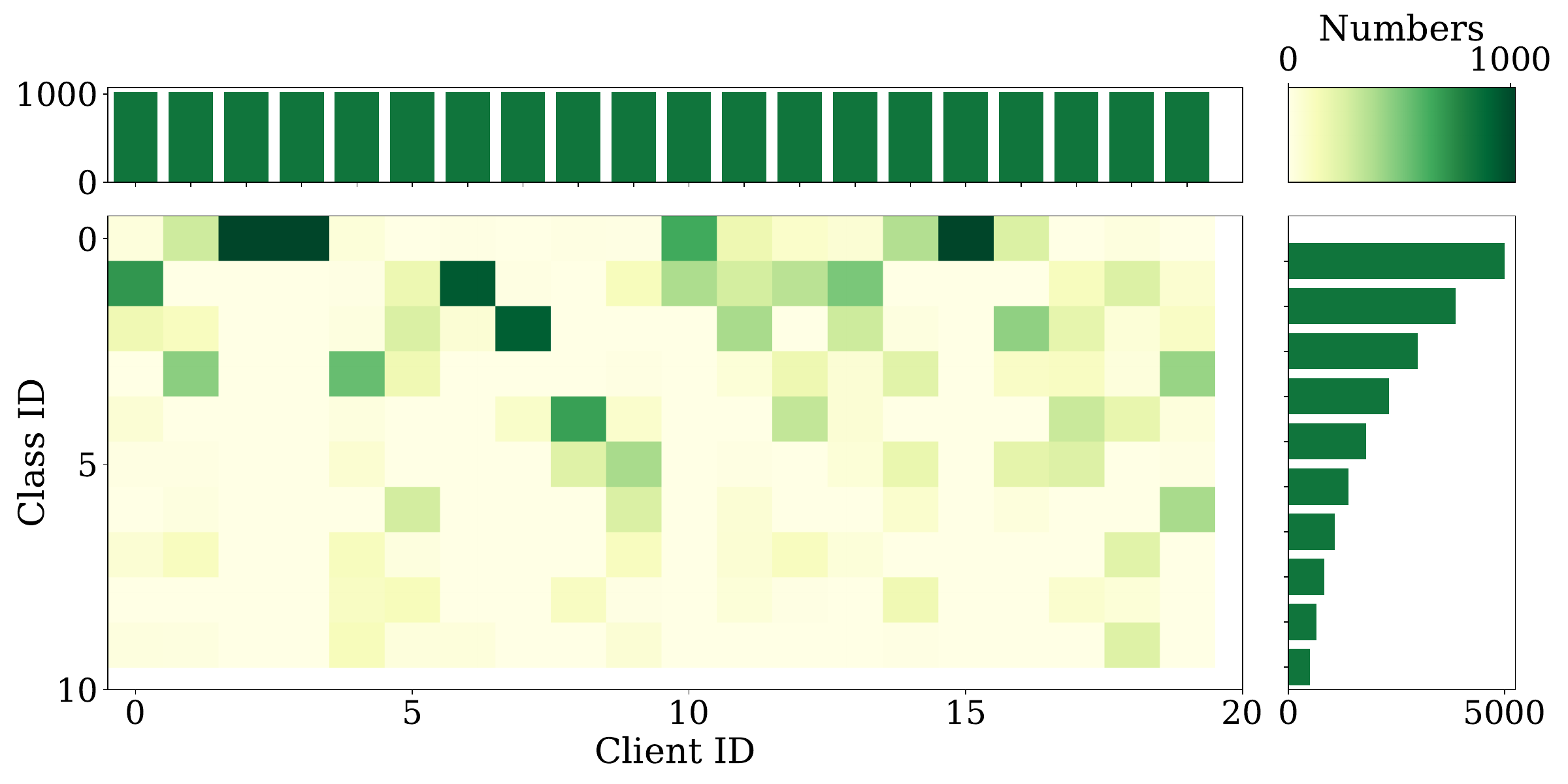}
    \caption{CIFAR10-LT, $\alpha=0.5$}
  \end{subfigure}
  \begin{subfigure}{\linewidth}
   \includegraphics[width=\linewidth]{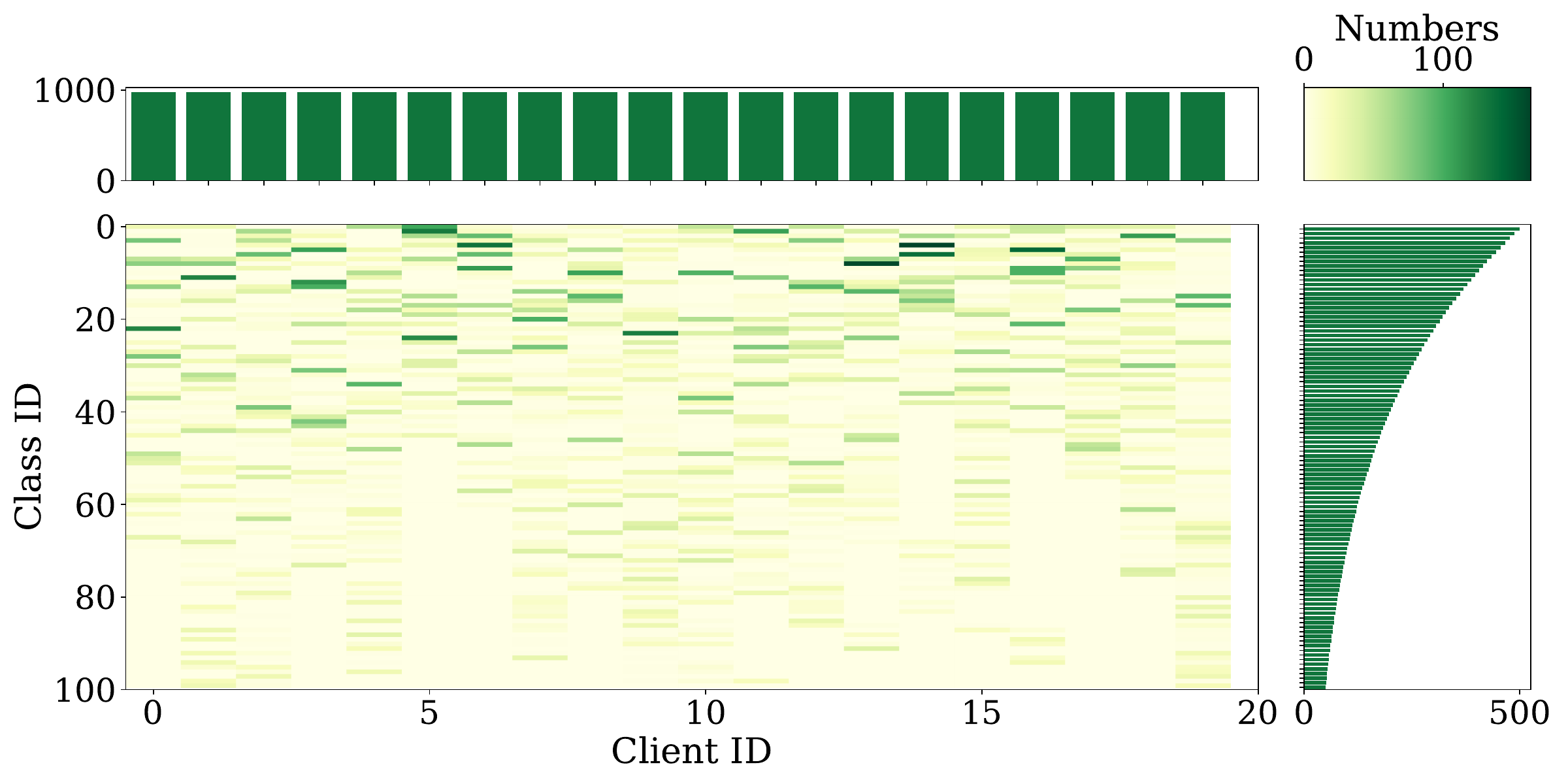}
    \caption{CIFAR100-LT, $\alpha=0.5$}
  \end{subfigure}
  \caption{The data distributions for CIFAR10-LT and CIFAR100-LT with IF = 100. The lower-left plot shows each client's data distribution across 20 clients. The upper-left plot displays sample counts per client, and the lower-right shows sample counts per class. The color bar in the upper right represents the distribution intensity across clients.}
  \label{fig:data-distribution}
\vspace{-0.8cm}
\end{figure}

\section{More Experimental Setup}
\label{data_show}

~\cref{fig:data-distribution} illustrates the data distributions of CIFAR10-LT and CIFAR100-LT with an imbalance factor (IF) of 100. Lighter colors indicate fewer samples, highlighting the sparsity of data distribution under the federated long-tailed setting.

\section{More Analysis of FedYoYo}
\label{fedyoyo_analsys}
\subsection{Global-to-local Model Gap}
\begin{figure}
  \centering
  \begin{subfigure}{\linewidth}
    \includegraphics[height = 5cm]{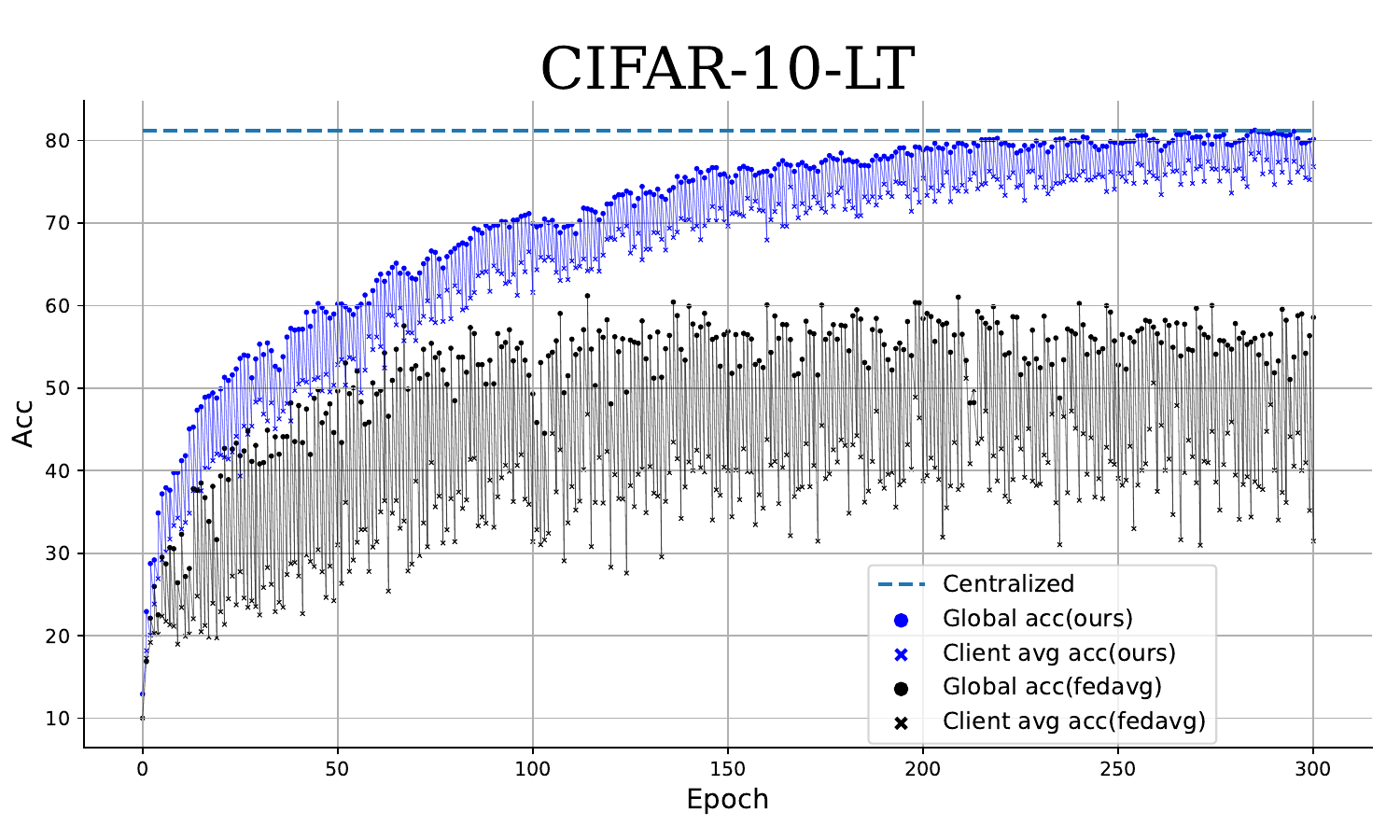}
    \caption{CIFAR-10-LT.}
  \end{subfigure}
  \begin{subfigure}{\linewidth}
   \includegraphics[height = 5cm]{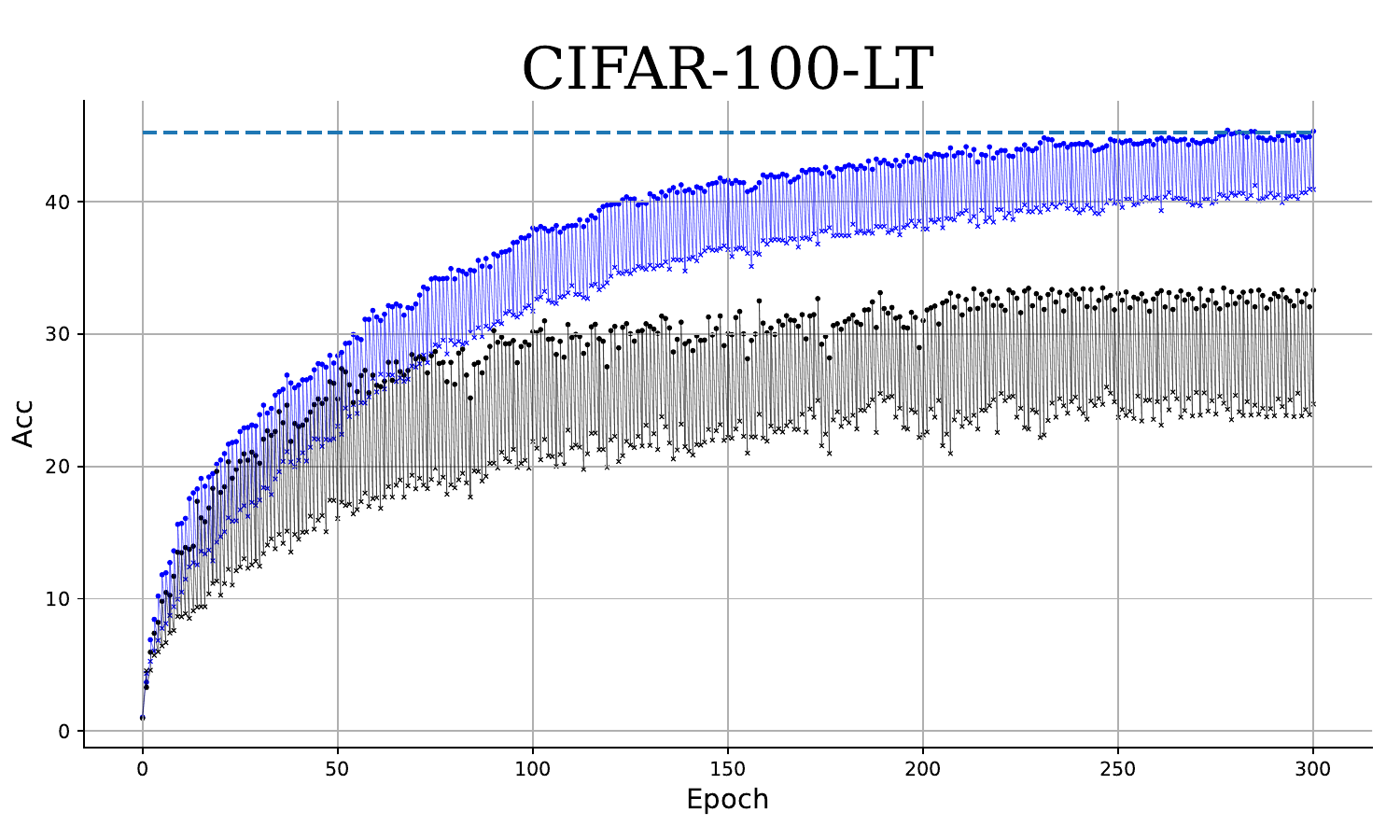}
    \caption{CIFAR-100-LT.}
  \end{subfigure}
  \caption{Comparison of global-client gap. }
  \label{fig:global-client-gap}
\end{figure}

As demonstrated in ~\cref{fig:global-client-gap}, our method effectively reduces the Global-to-local model gap in long-tailed data distribution scenarios, significantly enhancing the global model's accuracy. This improvement is especially pronounced on the CIFAR100-LT dataset, where our approach consistently outperforms the baseline methods and achieves higher convergence rates, thereby demonstrating superior generalization across clients.

\subsection{Feature Consistency}
\begin{figure}
  \centering
  \begin{subfigure}{0.49\linewidth}
    \includegraphics[width=\linewidth]{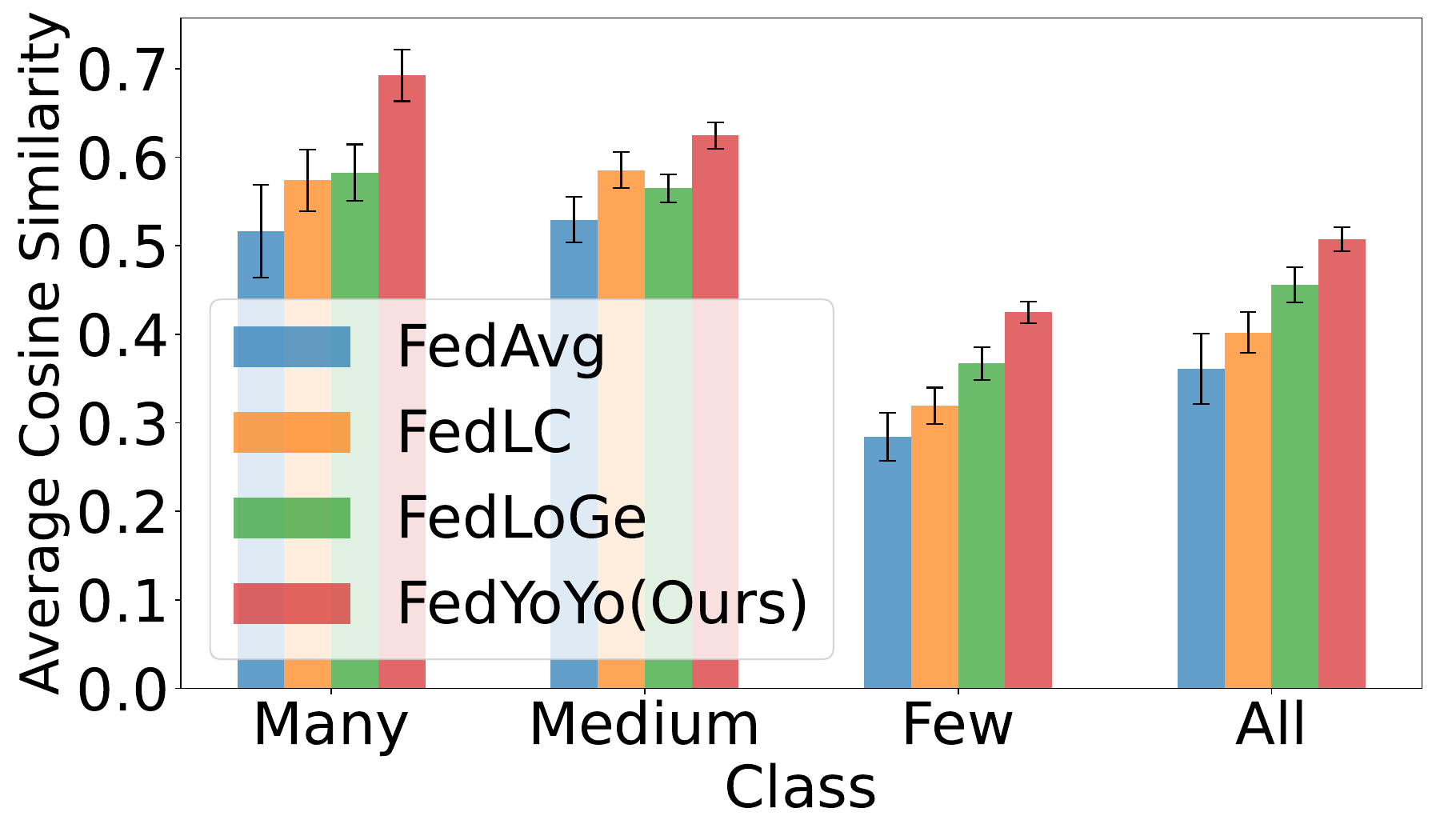}
    \caption{CIFAR-10-LT}
    \label{fig:cos_cifar10}
  \end{subfigure}
  \hfill
  \begin{subfigure}{0.49\linewidth}
    \includegraphics[width=\linewidth]{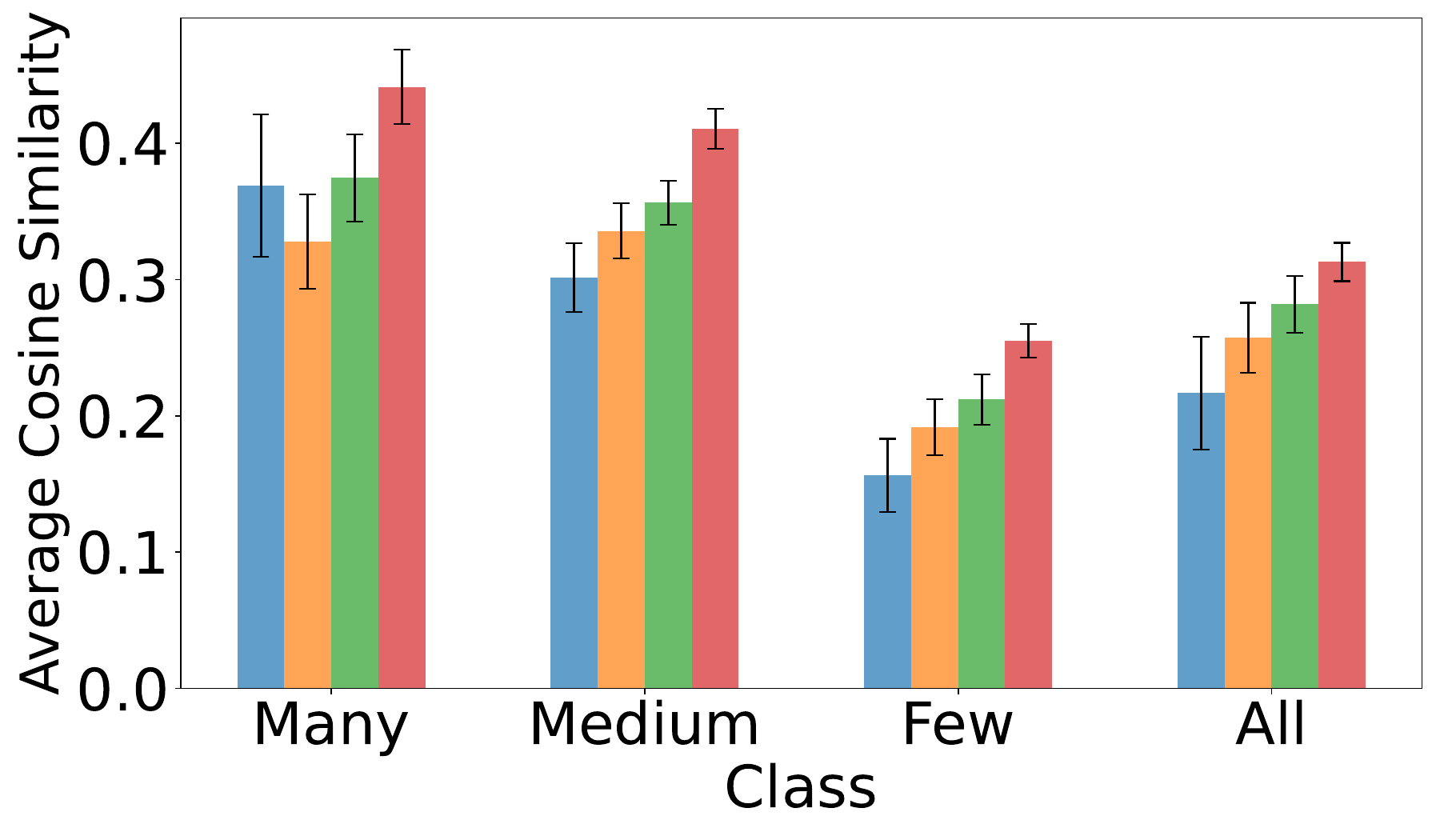}
    \caption{CIFAR-100-LT}
    \label{fig:cos_cifar100}
  \end{subfigure}
  % \vspace{-0.2cm}
  \caption{\textbf{Comparison of feature similarity between global model and local models.}}
  \label{fig:feature-consistency}
  \vspace{-0.1cm}
\end{figure}
In ~\cref{fig:feature-consistency}, we illustrate the average cosine similarity between local models and the global model across different class categories: Many, Medium, Few, and All. Our method consistently achieves the highest similarity on both CIFAR-10-LT and CIFAR-100-LT datasets. This higher similarity reflects a reduced discrepancy between local and global models, indicating that our approach effectively mitigates client drift. These findings confirm that our method can better align local models with the global model, thereby enhancing the overall consistency and performance in federated long-tailed learning scenarios.

\subsection{Effectiveness of Estimated Global Distribution}
\begin{figure}
  \centering
  \begin{subfigure}{0.49\linewidth}
    \includegraphics[width=\linewidth]{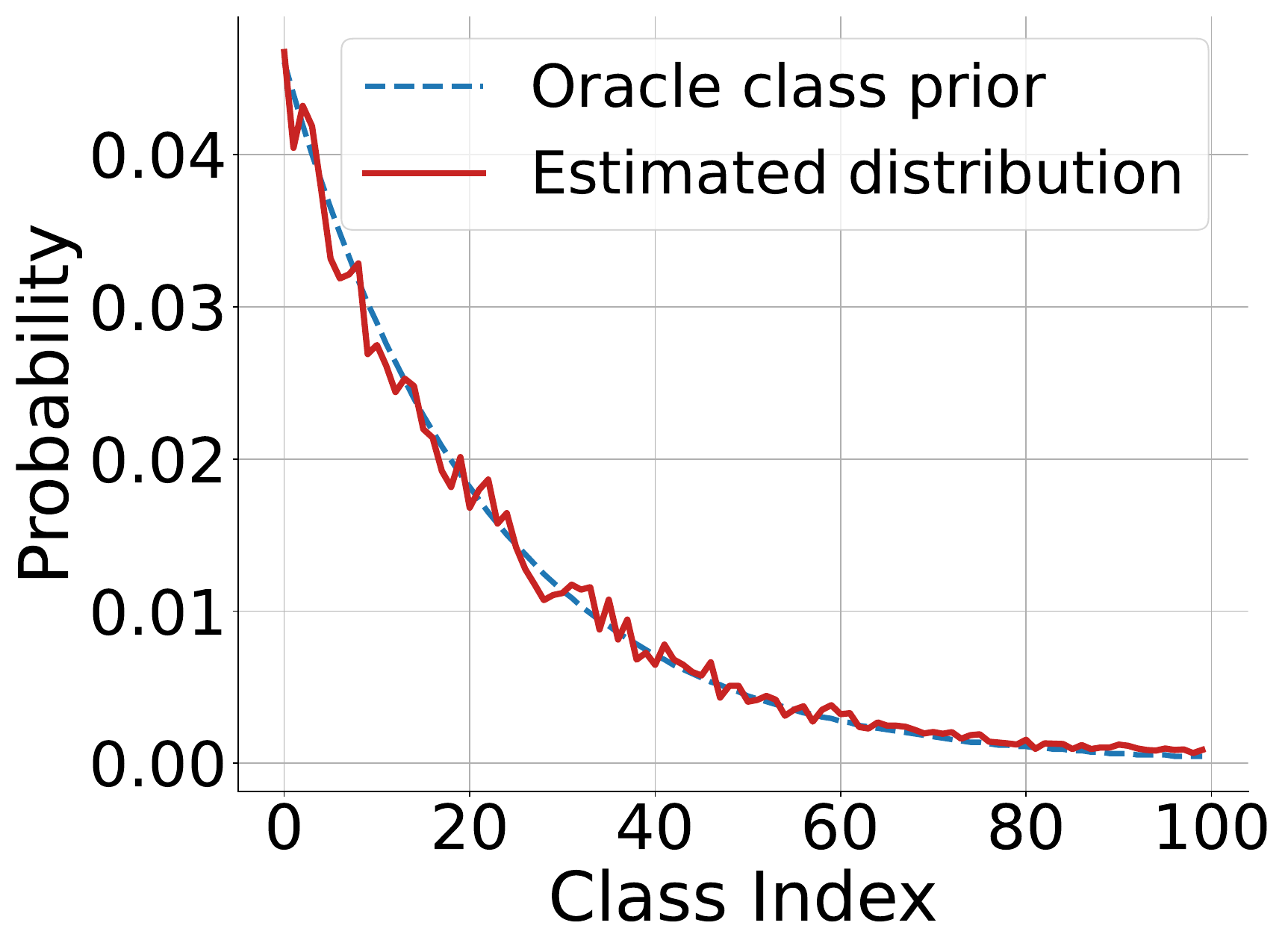}
    \caption{The estimated distribution.}
    \label{fig:class distribution}
  \end{subfigure}
   \hfill
  \begin{subfigure}{0.49\linewidth}
    \includegraphics[width=\linewidth]{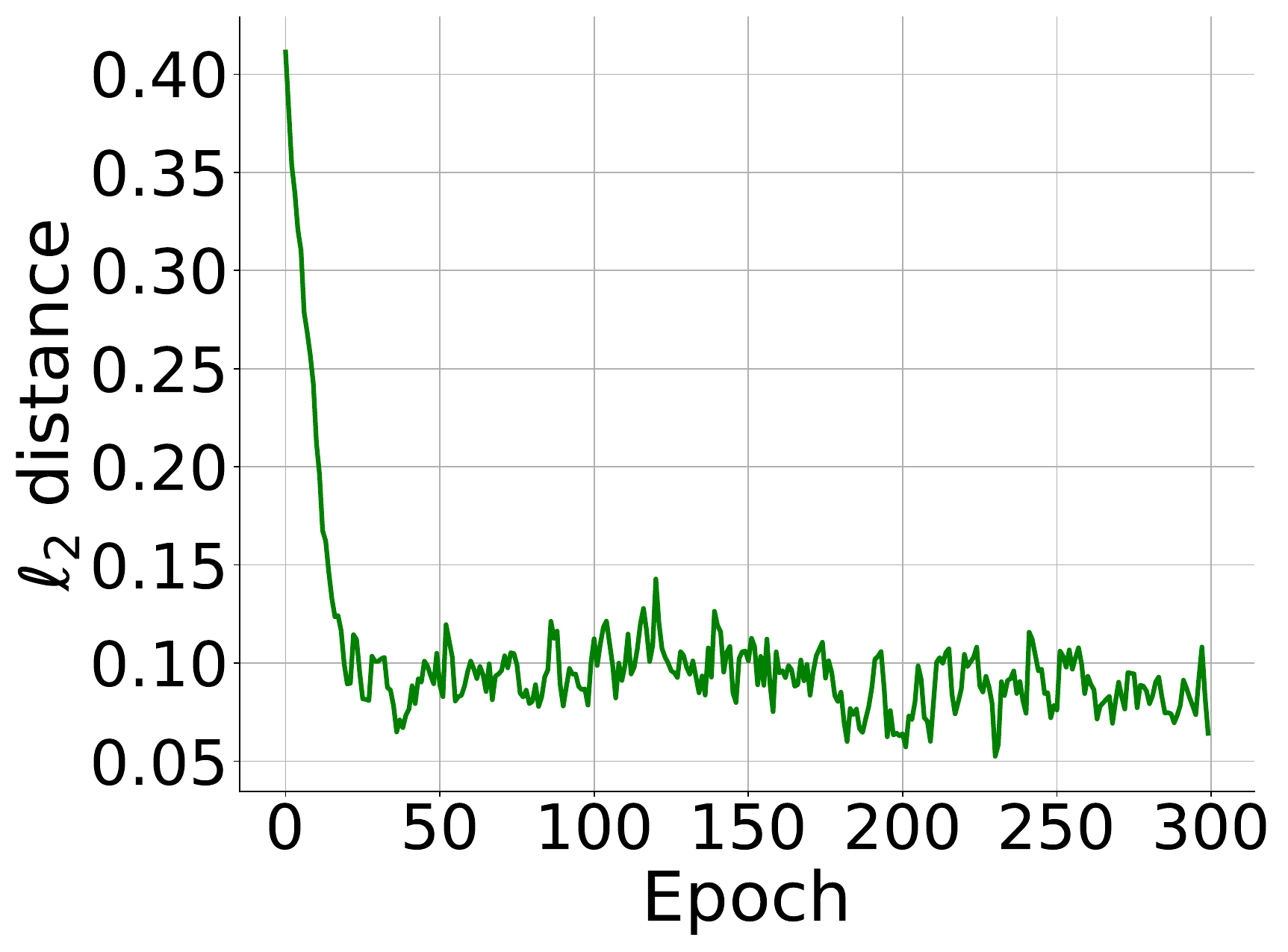}
    \caption{{$\ell_2$} distance during training.}
    \label{fig:l2_distance}
  \end{subfigure}
  \vspace{-0.2cm}
  \caption{\textbf{(a) The global long-tailed distribution obtained through estimated.} \textbf{(b) {$\ell_2$} distance between the statiscal distribution and the oracle class prior during training.} Experiments is conducted on CIFAR-100-LT with IF $= 100$ and {$\alpha=0.5$}. }
  \label{fig:dis-l2}
  \vspace{-0.2cm}
\end{figure}

 As shown in \cref{fig:l2_distance}, the distance steadily decreases, indicating that our parameterized class distribution converges toward the oracle prior over time. These results suggest that our model captures robust feature representations, as improved feature extraction directly enhances the accuracy of the distribution estimation, forming a mutually reinforcing cycle.

\section{More Ablation Experiments}
\label{more_ablation}
\begin{table}[t]
\begin{center}
\caption{\textbf{Top-1 test accuracy ({\%})  of different data augment policys.}}
% \vspace{-0.2cm}
\setlength\tabcolsep{8.0pt}
\scalebox{0.80}{
\begin{tabular}{ccccccc} 
\toprule
\multirow{2}{*}{Policys} & \multicolumn{3}{c}{CIFAR-10-LT} & \multicolumn{3}{c}{CIFAR-100-LT}  \\
% \cline{2-10}
\cmidrule(lr){2-7}
& RA & AA & TA & RA & AA & TA         \\
\midrule
Many    & 81.48	 & 81.95	& 82.92	& 60.26	& 59.6	& 58.66 \\
Medium	& 83.30	& 80.90	& 78.63	& 47.00	& 47.51	& 47.91 \\
Few	    & 79.57	& 81.20	& 80.60	& 28.63	& 28.4	& 29.37 \\
All 	& 81.45	& 81.41	& 80.94	& 46.13	& 46.01	& 46.11 \\
\bottomrule
\end{tabular}}
\label{tab:aug-policys}
\end{center}
\vspace{-0.5cm}
\end{table}

To evaluate the impact of different data augmentation strategies, we compare RandAugment (RA)~\cite{randaugment}, AutoAugment (AA)~\cite{autoaugment}, and TrivialAugment (TA)~\cite{trivialaugment}. As shown in ~\cref{tab:aug-policys}, RA achieves the best performance on both CIFAR-10-LT and CIFAR-100-LT, particularly in the Many and Medium class categories. However, the differences between these augmentation strategies are not significant, suggesting that the choice of augmentation method has limited impact on the overall performance. This indicates that the effectiveness of our method does not rely heavily on the specific augmentation strategy, but rather on the robustness of the approach itself.

\end{document}